\documentclass[10pt,twocolumn,letterpaper]{article}

\usepackage{cvpr}
\usepackage{times}
\usepackage{epsfig}
\usepackage{graphicx}
\usepackage{amsmath}
\usepackage{amssymb}
\usepackage{authblk}
\usepackage[symbol]{footmisc}
\renewcommand{\etc}{etc\@ifnextchar.{}{.\@}}

\renewcommand{\ie}{i.e., \@}
\renewcommand{\vs}{vs. \@}

\newcommand{\rb}{\right]}
\newcommand{\lb}{\left[}

\usepackage[breaklinks=true,bookmarks=false]{hyperref}

\cvprfinalcopy % *** Uncomment this line for the final submission
\def\cvprPaperID{****} % *** Enter the CVPR Paper ID here

\begin{document}

%%%%%%%%% TITLE
\title{Robust neural circuit reconstruction from serial electron microscopy with convolutional recurrent networks}

\author[1,$\dagger$]{Drew Linsley}
\author[1,$\dagger$]{Junkyung Kim}
\author[2]{David Berson}
\author[1]{Thomas Serre}
\affil[1]{Department of Cognitive, Linguistic \& Psychological Sciences}
\affil[2]{Carney Institute for Brain Science}
\affil[ ]{Brown University, Providence, RI \authorcr {\tt \{drew\_linsley, junkyung\_kim, david\_berson, thomas\_serre\}@brown.edu} \vspace{-2ex}}

% \author{Drew Linsley\\
% Department of Cognitive Linguistic & Psychological Sciences\\
% Carney Institute for Brain Science\\
% Brown University\\
% {\tt\small firstauthor@i1.org}
% \and
% Junkyung Kim\\
% Department of Cognitive Linguistic & Psychological Sciences\\
% Carney Institute for Brain Science\\
% Brown University\\
% {\tt\small firstauthor@i1.org}
% \and
% David Berson\\
% Department of Neuroscience\\
% Carney Institute for Brain Science\\
% Brown University\\
% {\tt\small firstauthor@i1.org}
% \and
% Thomas Serre\\
% Department of Cognitive Linguistic & Psychological Sciences\\
% Carney Institute for Brain Science\\
% Brown University\\
% {\tt\small firstauthor@i1.org}
% }

\maketitle
%\thispagestyle{empty}

%%%%%%%%% ABSTRACT
\begin{abstract}
Recent successes in deep learning have started to impact neuroscience. Of particular significance are claims that current segmentation algorithms achieve ``super-human'' accuracy in an area known as connectomics. However, as we will show, these algorithms do not effectively generalize beyond the particular source and brain tissues used for training -- severely limiting their usability by the broader neuroscience community. To fill this gap, we describe a novel connectomics challenge for source- and tissue-agnostic reconstruction of neurons (STAR), which favors broad generalization over fitting specific datasets. We first demonstrate that current state-of-the-art approaches to neuron segmentation perform poorly on the challenge. We further describe a novel convolutional recurrent neural network module that combines short-range horizontal connections within a processing stage and long-range top-down connections between stages. The resulting architecture establishes the state of the art on the STAR challenge and represents a significant step towards widely usable and fully-automated connectomics analysis.
\end{abstract}

%%%%%%%%% BODY TEXT
\section{Introduction}

\renewcommand{\thefootnote}{\fnsymbol{footnote}}
\footnotetext[2]{These authors contributed equally to this work.}
\renewcommand*{\thefootnote}{\arabic{footnote}}

\begin{figure*}[t]
\begin{center}
   \includegraphics[width=\textwidth]{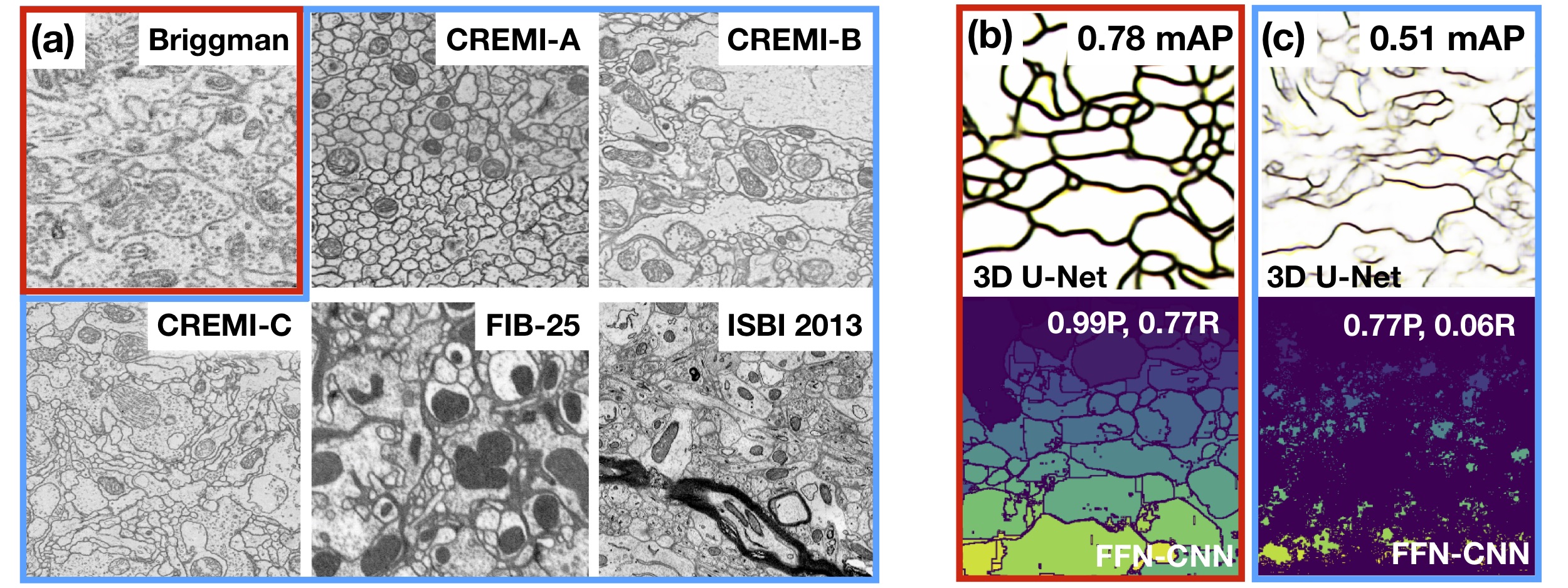}
\end{center}\vspace{-2mm}
   \caption{State-of-the-art deep learning architectures~\cite{Lee2017-ye,Januszewski2018-rl} for neuron reconstruction generalize to held-out images from the same SEM volume but not to different volumes. (a) Representative SEM images. Samples collected from the fruit fly nervous system (Volumes A, B, and C volumes from CREMI, and FIB-25), mouse cortex (SNEMI3D), and retina (Briggman~\cite{Briggman2011-aw}). Colored boxes indicate the dataset used to train the models that produced the results in (b) and (c). (b) The top-panel shows performance of a state-of-the-art 3D U-Net~\cite{Lee2017-ye} trained and tested on distinct sections of the Briggman volume (red outline in a) to predict the locations of membranes. A leading approach to neuron reconstruction utilizes these predictions as scaffold for segmenting neurons with post-processing methods like watershed. The 3D U-Net makes high-quality predictions of the locations of membranes, as measured by the contour detection metric mean average precision (mAP~\cite{Xiaofeng2012-ho}; calculated per-slice and then averaged; higher is better). An alternate approach to neuron reconstruction is the Flood-filling network (FFN)~\cite{Januszewski2018-rl}, which successfully segments the Briggman volume when trained and tested on distinct sections of it (lower panel). Its performance is measured by precision and recall (P, R; ~\cite{Haehn2017-ma}; higher is better). (c) The STAR challenge involves training models on images from multiple datasets (blue outline in a), and test on images from an independent dataset (Briggman). This challenge severely strains membrane predictions of the 3D U-Net (top-panel of c) and segmentations from the FFN (bottom-panel of c), demonstrating their limited generalization.}
\label{fig:teaser}
\end{figure*}

Connectomics -- the comprehensive mapping of synaptic circuits -- is poised to revolutionize our understanding of neural computations. Technical advances permit generation of high-resolution serial electron microscope (SEM) images spanning volumes hundreds of microns across. These can reveal synaptic associations among classes of neurons and advance our understanding of the circuits underlying specific neural computations. Analysis software and crowd-sourcing platforms are being developed to mine these invaluable and extremely large data sets, and initial results are impressive (see~\cite{Fornito2015-tt, Swanson2016-ls, Van_den_Heuvel2016-if, Schroter2017-qm} for reviews.) 

For example, SEM imaging of the retina was used to explain the direction-selectivity of certain ganglion cells based on their asymmetric contacts with a particular type of inhibitory interneuron~\cite{Takemura2013-ch}. Various labs are spearheading applications in the thalamus and visual cortex of the mouse, as well as the fruit fly nervous system, among other model organisms~\cite{Vishwanathan2017-ta, Kim2014-bc, Briggman2011-aw, Lee2017-ye, Ding2016-ui, Sabbah2017-jn}. Yet despite its great promise, SEM has not gained widespread adoption among neuroscientists. 

One of the key steps in connectomics involves segmenting neurons from their surroundings -- a task that requires such time intensive manual annotation that it is in effect impossible for all but the most basic organisms~\cite{White1986-kv}. Computer vision has been critical to improving the efficiency of neuron segmentation and enabling the reconstruction of complex nervous systems at ``superhuman'' levels of accuracy~\cite{Lee2017-ye}. These successes hint at the possibility of automating neuron reconstruction in SEM volumes at scales that would be unimaginable for manual work. As an example, it has been reported that computer vision permits a 250-fold reduction in processing time over brute-force manual annotation for a 100$^{3}$-$\mu m^{3}$ tissue volume~\cite{Januszewski2018-rl}. However, the 400 man-hours of expert annotation required for the generation of this system's training data still represents a significant barrier. For computer vision to be most useful in practice, segmentation algorithms must generalize across brain tissues to allow completely automated mapping of neural circuits.

Consider Fig.~\ref{fig:teaser}a, which depicts sample SEM images from a variety of connectomics challenges~\cite{Briggman2011-aw,Kasthuri2015-fh}. Current state-of-the-art approaches~\cite{Lee2017-ye,Januszewski2018-rl} are trained and tested on disjoint sections of the same volumes (Fig.~\ref{fig:teaser}b). However, we have found that when asked to make judgments on different tissues (Fig.~\ref{fig:teaser}c), these network architectures do not generalize well. Hence, model weights successfully trained on one dataset must be ``fine-tuned'' on tissue-specific annotations from another dataset to make accurate predictions on those images. Most neurobiology laboratories do not have the resources to annotate and curate SEM datasets at this scale, and they do not have the computer vision expertise to retrain state-of-the-art computer vision systems for their specific brain volumes. The development of computer vision algorithms that can generalize across brain volumes represents the next frontier in automating neuron reconstructions.

% A more desirable computer vision system for connectomics is one that can accurately segment neurons without needing to be trained on the idiosyncratic qualities of the cellular composition of a particular tissue and imaging apparatus.  

\vspace{-1mm}\paragraph{Contributions} Here, we address the poor generalization of computer vision systems in connectomics with a novel challenge: the source- and tissue-agnostic reconstruction (STAR; \url{STAR-challenge.github.io}). This challenge presents a ``training'' dataset consisting of five publicly available and annotated volumes of SEM images  representing a variety of organisms and imaging configurations (CREMI, FIB-25, and SNEMI3D; see Table~\ref{tab:datasets}); evaluation is performed on an independent volume~\cite{Briggman2011-aw} annotated by our group. We will demonstrate that state-of-the-art systems for neuron reconstruction fail on the STAR challenge.

% \enlargethispage{2mm}
An alternative to the feedforward convolutional neural networks which form the backbone of state-of-the-art segmentation architectures can be inferred from the organization of the visual cortex, where feedback connections between neurons mediate their highly complex interactions and are thought to form the basis for robust vision. Recent work motivated by biological considerations introduced the horizontal gated recurrent unit (hGRU), and found that such horizontal interactions between computational units support parameter-efficient solutions to contour detection tasks~\cite{Linsley2018-ls}.

We build on this framework and introduce the feedback gated recurrent unit (fGRU), which generalizes the recurrent connections introduced in the hGRU to include not only horizontal connections within a layer but also top-down connections between processing layers. We find that fGRU models perform on par or better than state-of-the-art systems for neuron reconstruction when trained and tested on tissue images sampled from the same volumes. Importantly, fGRU models also significantly outperform these systems on the STAR challenge, where generalizing to an independent volume is needed. %More generally, the present study offers a path forward for completely automated neural circuit reconstruction.

% \enlargethispage{2mm}

\section{Related work}

\paragraph{Neuron reconstruction datasets}
The development of computer vision systems for connectomics has been largely supported by the public availability of SEM datasets paired with expert annotations. These include volumes sampled from fruit fly nervous system (CREMI, \url{cremi.org}; FIB-25, \url{http://janelia-flyem.github.io/}) and mouse visual cortex (SNEMI3D, \url{http://brainiac2.mit.edu/SNEMI3D/}). The proposed STAR challenge combines these datasets with one from mouse retina~\cite{Briggman2011-aw}. Each of the volumes was fixed and imaged using slightly different methods, but at least to the human eye, appear visually similar (Fig.~\ref{fig:teaser}; see Table~\ref{tab:datasets} for dataset details).

 \paragraph{Automated 3D neuron reconstruction}
One classic computer vision pipeline to neuron reconstruction involves the following three stages: First, cell membranes are detected; second, a watershed algorithm grows supervoxels within each cell membrane; third, an additional post-processing step agglomerates supervoxels into more accurate segmentation masks~\cite{Nunez-Iglesias2013-eo}. One of the leading architectures featuring such an approach uses a 3D U-Net architecture for membrane detection -- resulting in ``superhuman'' segmentation accuracy in the SNEMI3D challenge~\cite{Lee2017-ye}. This network architecture combines the encoder-decoder structure of a typical U-Net~\cite{Ronneberger2015-ru} with extensive ``residual connections''~\cite{He2015-lm} that mix information between different processing stages, and losses for predicting voxel affinities.

A more recent approach integrates membrane detection and segmentation into the same end-to-end trainable neural network architecture~\cite{Januszewski2018-rl}. Flood-filling networks (FFNs) segment volumes one neuron at a time. To date, they have outperformed all other approaches in processing the FIB-25 dataset. The core of this model is a 3D CNN with residual connections, which receives two types of inputs: a 3D volume of SEM images, and a corresponding mask volume which is iteratively updated by the network to assign voxels from the volume to a neuron under consideration. In the first iteration, the network processes a small 3D crop (which is called the network's ``field of view'') of the volume centered around a randomly initialized X/Y/Z ``seed'' location. Updates to the mask volume inform the model where to move its field of view on the next processing step. This process continues until an entire cell has been labeled.

%%%
\begin{table*}[t!]
\centering
\begin{tabular}{|c|c|c|c|c|}
\hline Name & Tissue & Imaging & Resolution & Voxels (X/Y/Z/Volumes) \\ \hline
CREMI & Fruit fly & ssTEM~\cite{Stevens1980-ef} & $4\times4\times40$nm & $1250\times1250\times 125\times3$\\
FIB-25 & Fruit fly & FIBSEM~\cite{Kizilyaprak2014-fs} & $8\times8\times8$nm & $520\times520\times520\times1$ \\
SNEMI3D~\cite{Kasthuri2015-fh} & Mouse cortex & mbSEM~\cite{Eberle2015-cq} & $6\times6\times29$nm & $1024\times1024\times100\times1$ \\
Briggman~\cite{Briggman2011-aw} (Ours) & Mouse retina & SBEM~\cite{Denk2004-zc} & $16.5\times16.5\times23$nm &  $384\times384\times384\times1$ \\\hline
\end{tabular}
\linebreak\linebreak%\vspace{-4mm}
\caption{SEM image volumes used in the STAR challenge. The training dataset consists of CREMI, FIB-25, and SNEMI3D, whereas the test dataset, Briggman, is a volume from~\cite{Briggman2011-aw} annotated by us.}
\label{tab:datasets}%\vspace{-5mm}
\end{table*}
%%%%

\paragraph{Recurrent vision models}
Recurrent neural networks (RNNs) have classically been used for modeling temporal sequences, but work over recent years has started to demonstrate their utility for general vision tasks. Successful RNN applications in vision include object recognition and super-resolution tasks, where convolution kernels are applied recursively to increase processing depth without adding parameters~\cite{Liang2015-li,Liao2016-me,Kim2016-pg}. A growing body of work has shown the benefits of introducing learning rules or connectivity patterns that are inspired by the anatomy and/or the physiology of the visual cortex into RNNs~\cite{Spoerer2017-ee,Lotter2016-qr,Zamir2016-lr,Nayebi2018-dc,Linsley2018-ls}. 

In particular, the horizontal gated recurrent unit (hGRU) aims to approximate a recurrent-network neuroscience model of contextual illusions~\cite{Mely2018-bc}, with a convolutional-RNN module that can be trained using routines from deep learning. The resulting module uses trainable gates, borrowed from the gated recurrent unit (GRU~\cite{Cho2014-tn}), to control the integration of information over time. A single hGRU layer learned to solve contour detection tasks more efficiently than CNNs (with many times the number of parameters) by learning patterns of horizontal connectivity between units at different spatial locations and across feature channels. 

The proposed feedback gated recurrent unit (fGRU) builds on the hGRU, extending it into a general-purpose module for learning feedback interactions -- both short- (\ie horizontal connections within a layer) and long-range (\ie top-down connections between layers).

%  \paragraph{Push-pull neural models}
%~\cite{Grossberg1985-ui,Series2003-mg,Zhaoping2011-dn,Shushruth2012-dv,Rubin2015-ws, Mely2018-bc}
% % - what are the push-pull dynamics and what do they do for us
% % - mention gilbert association field feedback
% % - discuss how perceptual grouping is a good inductive bias for segmentation tasks

\section{Feedback gated recurrent unit}
\paragraph{Recurrent-network neuroscience model}

% Computational neuroscience models of feedback in visual cortex have demonstrated how perceptual grouping-like phenomena, such as contour integration, can emerge from interactions between units within and between processing layers~\cite{Grossberg1985-ui,Series2003-mg,Zhaoping2011-dn,Shushruth2012-dv,Rubin2015-ws, Mely2018-bc}. It has recently been shown that the dynamics of these models can be incorporated into an end-to-end trainable module, the hGRU~\cite{Linsley2018-ls}. 

We briefly review the hGRU~\cite{Linsley2018-ls}, which this work builds upon. The hGRU implements the dynamics of a recurrent-network model of horizontal interactions~\cite{Mely2018-bc}, which explained human psychophysical responses to contextual illusions. We begin by describing this model. Its units are indexed by their 2D position $(x, y)$ and feature channel $k$. Neural activity is governed by the following RNN, which approximates Euler integration on the dynamical system described in~\cite{Mely2018-bc}:
% , beginning with a model of push-pull interactions between neurons that implements basic cues for perceptual grouping, then describing a version in which the parameters governing the dynamics as well as the connections between units can be trained with gradient descent.
% The model of horizontal connections in cortex developed by~\cite{Mely2018-bc} explained human psychophysical responses to contextual illusions, such as the orientation-tilt effect [REF]. A version of this model that can be trained with gradient descent was derived by~\cite{Linsley2018-ls}. 
% \enlargethispage{2mm}
\vspace{-1mm}\begin{align}
\begin{split}
\textbf{C}^{I} &= (\textbf{W}^{I} * \textbf{H}^{(2)}[t-1])\\
\textbf{C}^{E} &= (\textbf{W}^{E} * \textbf{H}^{(1)}[t])\\
H_{xyk}^{(1)}[t]&=\epsilon^{-2}\lb \xi X_{xyk} - (\alpha H_{xyk}^{(1)}[t-1] + \mu)C_{xyk}^{I}[t]\rb_+\\
H_{xyk}^{(2)}[t]&=\epsilon^{-2}\lb\gamma C_{xyk}^{E}[t]\rb_+.\label{euler}\\
% \mbox{where},\\
% \end{align}
% \vspace{-1mm}\begin{align*}
% C^{(1)}_{xyk} &= [W * H^{(2)}]_{xyk}\\
% C^{(2)}_{xyk} &= [W * H^{(1)}]_{xyk},
\end{split}%\vspace{-2mm}
\end{align}

These equations describe the evolution of recurrent synaptic inputs and outputs $\textbf{H}^{(1)}, \textbf{H}^{(2)} \in \mathbb{R}^{\textit{W} \times \textit{H} \times \textit{K}}$, which are influenced by the feedforward drive $\textbf{X} \in \mathbb{R}^{\textit{W} \times \textit{H} \times \textit{K}}$ (\ie the neural response to a stimulus) over discrete timesteps, denoted by $\cdot[t]$. By separating contributions from synaptic inputs and outputs, the model is able to simulate different known forms of linear and non-linear inhibitory and excitatory interactions between neurons (see~\cite{Mely2018-bc} for details), implemented by the kernels $\textbf{W}^{I}$ and $\textbf{W}^{E}$ on the on the input and output, respectively. 

The activity $\textbf{C}^{I}$ describes horizontal inhibitory interactions on neuron outputs, $\textbf{H}^{(2)}$. Inhibition is a function of both the output, via the $\textbf{C}^{I}$ activity, and the input, via the $\textbf{C}^{E}$ activity. This supports both linear and non-linear forms of inhibition; the latter of which is referred to as ``shunting'' (or divisive) inhibition~\cite{Grossberg1985-ui}. Shunting inhibition means that the effect of the inhibition on the neuron will scale with its output activity (the more active the neuron the stronger the inhibition will be). On the other hand, excitation in the model is a linear function of $\textbf{C}^{E}$, which describes horizontal excitatory interactions between units in $\textbf{H}^{(1)}$. Linear and non-linear forms of inhibition are controlled by the parameters $\mu$ and $\alpha$, respectively, whereas linear excitation is controlled by $\gamma$. The ReLU function is denoted by $[\cdot]_+=\max(\cdot,0)$, and $\epsilon$ is a time constant.

\begin{figure}[t!]
\begin{center}
   \includegraphics[width=.98\linewidth]{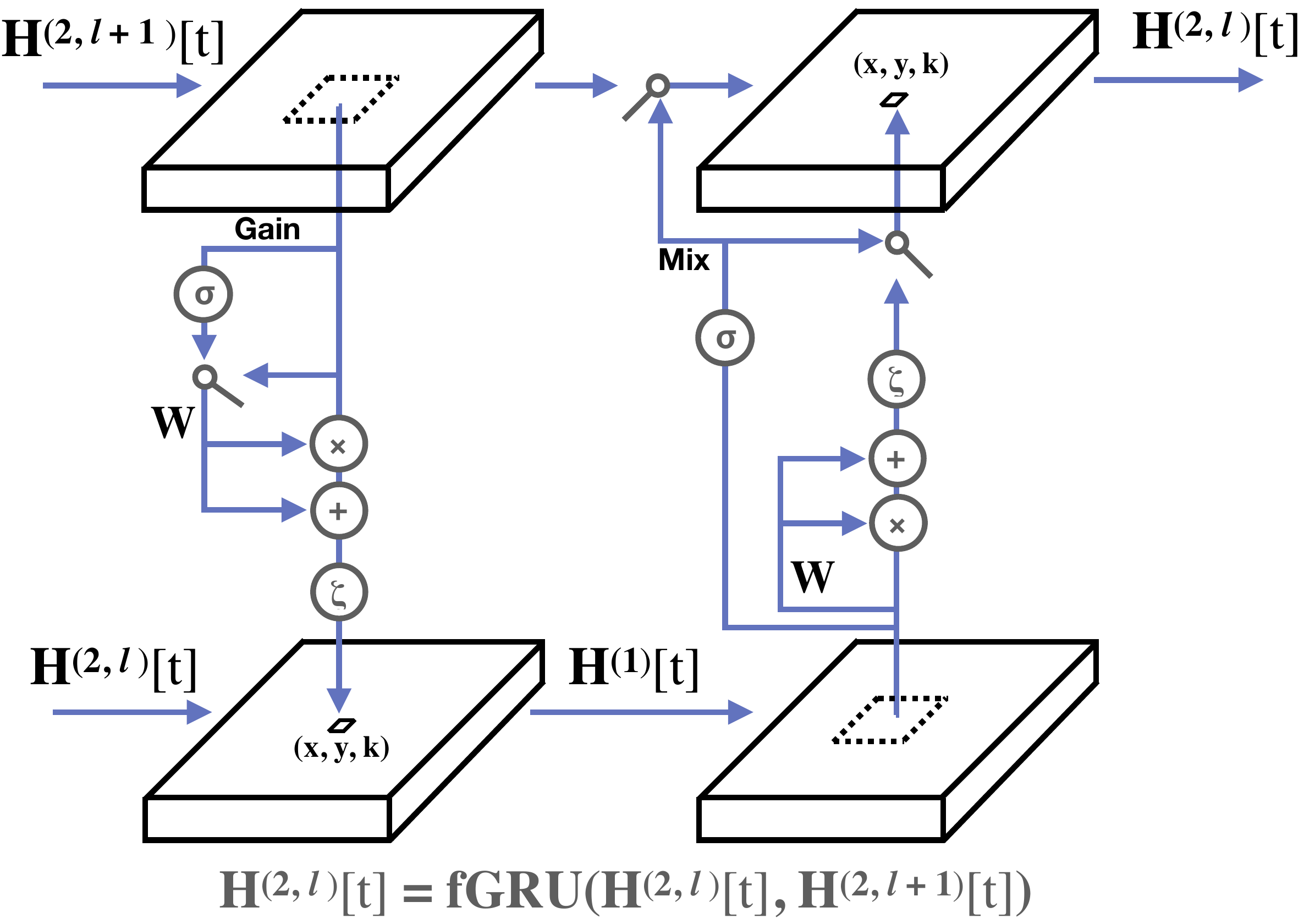}
\end{center}%\vspace{-4mm}
   \caption{A diagram of the top-down module of the fGRU, which is a generalization of the hGRU~\cite{Linsley2018-ls} that allows for horizontal or top-down connectivity. The fGRU shares the same internal dynamics as the hGRU (captured by the function fGRU in the equation below the figure) but the input and output activities (the feedforward drive \textbf{X} and the persistent output activity $\textbf{H}^{(2)}[t-1]$) in the hGRU are replaced by $\textbf{H}^{(2,l)}[t]$ (the persistent output at layer $l$) and $\textbf{H}^{(2,l+1)}[t]$ (the persistent output at layer $l + 1$), respectively. In the first stage, a gain is applied to $\textbf{H}^{(2,l+1)}[t]$, and the resulting activity is convolved with the kernel $\textbf{W}$ to compute top-down interactions between units in layer $l+1$ and $l$: both linear ($+$ symbol) and quadratic ($\times$ symbol). In the second stage, another set of linear and quadratic operations modulate an activity derived from convolving the updated $\textbf{H}^{(1)}[t]$ with $\textbf{W}$. A mix calculates the fGRU output by integrating this candidate output with $\textbf{H}^{(2,l+1)}[t]$. Small solid-line squares within the circuit's hypothetical activities (big boxes) denote the unit indexed by 2D position $(x, y)$ and feature channel $k$. Dotted-line squares depict this unit's projective field (a union of both classical and extra-classical definitions). Sigmoid and hyperbolic tangent point nonlinearities are denoted by $\sigma$ and $\zeta$, respectively.}
\label{fig:arch_diagram}
\end{figure}

% \newpage
\paragraph{The hGRU module}\label{hgru}
The hGRU (Fig.~\ref{fig:arch_diagram}) extends the convolutional recurrent model described in Eq.~\ref{euler} with learnable gates, which facilitates learning with gradient descent~\cite{Linsley2018-ls}. We describe a formulation of this model that includes recurrent batch-norm~\cite{Cooijmans2016-yo}, which leads to more consistent training. This hGRU is governed by the following equations, which relax many of the constraints of Eq.~\ref{euler} that were shown to be unnecessary for contour detection tasks (see~\cite{Linsley2018-ls} for details):
\vspace{-1mm}\begin{align}
\begin{split}
    \textbf{G}^{(1)}[t] &= \sigma(\mathrm{BN}(\textbf{U}^{(1)} * \textbf{H}^{(2)}[t-1]))\\
    \textbf{C}^{(1)}[t] &= \mathrm{BN}(\textbf{W} * (\textbf{G}^{(1)}[t] \odot \textbf{H}^{(2)}[t-1]))\\
    \textbf{G}^{(2)}[t] &= \sigma(\mathrm{BN}(\textbf{U}^{(2)} * \textbf{H}^{(1)}[t])))\\
    \textbf{C}^{(2)}[t] &= \mathrm{BN}(\textbf{W} * \textbf{H}^{(1)}[t])\\
    {H}^{(1)}_{xyk}[t] &= \zeta({X}_{xyk} - {C}^{(1)}_{xyk}[t](\alpha_{k} {H}^{(2)}_{xyk}[t-1] +\mu_{k})) \\
    \tilde{{H}}^{(2)}_{xyk}[t] &= \zeta({\tau}_{k}{H}^{(1)}_{xyk}[t] + {\beta}_{k}{C}^{(2)}_{xyk}[t] + {\gamma}_{k}{H}^{(1)}_{xyk}[t]{C}^{(2)}_{xyk}[t])\\
    {H}^{(2)}_{xyk}[t] &= {H}^{(2)}_{xyk}[t-1]({1} - {G}^{(2)}_{xyk}[t]) + \tilde{{H}}^{(2)}_{xyk}[t]{G}^{(2)}_{xyk}[t]\\
    \mbox{where } &
    % \vspace{-1mm}\begin{align*}
    % C^{(1)}_{xyk} &= [W * H^{(2)}]_{xyk}\\
    % C^{(2)}_{xyk} &= [W * H^{(1)}]_{xyk},
    \mathrm { BN } ( \mathbf { r } ; \boldsymbol{\delta} , \boldsymbol{\nu} ) = \boldsymbol{\nu} + \boldsymbol{\delta} \odot \frac { \mathbf { r } - \widehat { \mathbb { E } } [ \mathbf { r } ] } { \sqrt { \widehat { \operatorname { Var } } [ \mathbf { r } ] + \eta } }.\label{hGRU}
\end{split}
\end{align}

%\begin{align}
%\begin{split}

%    {G}^{(1)}_{xyk}[t] &= \sigma((U^{(1)} * {H}^{(2)}[t-1])_{xyk} + {b}^{(1)}_{k})\\
%    {H}^{(1)}_{xyk}[t] &= \zeta({X}_{xyk} - {C}^{(1)}_{xyk}[t](\alpha_{k} {H}^{(2)}_{xyk}[t-1]\\
%    {H}^{(2)}_{xyk}[t] &= {H}^{(2)}_{xyk}[t-1]({1} - {G}^{(2)}_{xyk}[t]) + \tilde{{H}}^{(2)}_{xyk}[t]{G}^{(2)}_{xyk}[t] +\mu_{k})) \\
%     \tilde{{H}}^{(2)}_{xyk}[t] &= \zeta({\tau}_{k}{H}^{(1)}_{xyk}[t] + {\beta}_{k}{C}^{(2)}_{xyk}[t] + {\gamma}_{k}{H}^{(1)}_{xyk}[t]{C}^{(2)}_{xyk}[t])\\
%     {C}^{(1)}_{xyk}[t] &= \mathrm{BN}(\textbf{W} *   (\textbf{G}^{(1)}[t] \odot \textbf{H}^{(2)}[t-1]))_{xyk}\\
%     {C}^{(2)}_{xyk}[t] &= \mathrm{BN}(\textbf{W} * \textbf{H}^{(1)}[t])_{xyk}\\
%  G^{(1)}_{xyk}[t] &= \sigma(\mathrm{BN}(\textbf{U}^{(1)} * \textbf{H}^{(2)}[t-1])_{xyk})\\
    % \label{inhibition} \\
% \end{split}
% \end{align}
% \begin{align}\vspace{-1mm}
% \begin{split}
    % {G}^{(2)}_{xyk}[t] &= \sigma({X}_{xyk} + (U^{(2)} * {H}^{(1)}[t])_{xyk} + {b}_{k}^{(2)})\\
%    {G}^{(2)}_{xyk}[t] &= \sigma(\mathrm{BN}(\textbf{U}^{(2)} * \textbf{H}^{(1)}[t]))_{xyk})\\
% \end{align}%\vspace{-5mm}
%\mbox{where } &
% \vspace{-1mm}\begin{align*}
% C^{(1)}_{xyk} &= [W * H^{(2)}]_{xyk}\\
% C^{(2)}_{xyk} &= [W * H^{(1)}]_{xyk},
%\mathrm { BN } ( \mathbf { h } ; \delta , \nu ) = \nu + \delta \odot \frac { \mathbf { h } - \widehat { \mathbb { E } } [ \mathbf { h } ] } { \sqrt { \widehat { \operatorname { Var } } [ \mathbf { h } ] + \epsilon } }.\label{hGRU}
% \end{split}\end{align}

Here, the feedforward drive $\textbf{X}$ may correspond to the activity from a preceding convolutional layer.  As in the model described in Eq.~\ref{euler}, this hGRU contains input and output states ($\textbf{H}^{(1)}$ and $\textbf{H}^{(2)}$), that allow it to capture complex nonlinear interactions between units in $\textbf{X}$. Updates to these hidden states are controlled by two activities: the ``gain'' $\textbf{G}^{(1)}[t]$, which modulates channels in $\textbf{H}^{(2)}[t-1]$ on every processing timestep, and the ``mix'' $\textbf{G}^{(2)}[t]$, which mixes a candidate $\tilde{\textbf{H}}^{(2)}[t]$ with the persistent $\textbf{H}^{(2)}[t]$. Both the gain and mix are transformed into the range $[0, 1]$ by a sigmoid nonlinearity ($\sigma$). Horizontal interactions between input units are computed in $\textbf{C}^{(1)}[t]$ by convolving $\textbf{H}^{(2)}[t]$ with the horizontal connection kernel $\textbf{W}$. Note that unlike in the original model from Eq.~\ref{euler}, the hGRU no longer implements separate kernels for excitatory and inhibitory connections.  Like Eq.~\ref{euler}, linear and quadratic interactions are controlled by the $k$-dimensional parameters $\boldsymbol{\mu}$ and $\boldsymbol{\alpha}$ (the latter of these corresponds to shunting inhibition described in Eq.~\ref{euler}). The pointwise function $\zeta$ is the squashing hyperbolic tangent that transforms activity into the range $[-1, 1]$. Separate applications of batch-norm are used on every timestep, where $\textbf{r} \in \mathbb { R } ^ { d }$ is the vector of batch-normalized preactivations. The parameters $\boldsymbol{\delta}$, $\boldsymbol{\nu} \in \mathbb { R } ^ { d }$ control the scale and bias of normalized activities, $\eta$ is a regularization hyperparameter, and $\odot$ is the elementwise multiplication.

Next, $\textbf{H}^{(1)}[t]$ is convolved with the kernel $\textbf{W}$ to compute $\textbf{C}^{(2)}[t]$, which captures interactions between neighboring units in the newly-computed $\textbf{H}^{(1)}[t]$. A candidate output $\tilde{\textbf{H}}^{(2)}[t]$ is calculated via the $k$-dimensional parameters $\boldsymbol{\tau}$ and $\boldsymbol{\beta}$, which control the linear terms of this interaction, and $\boldsymbol{\gamma}$ which controls the quadratic contributions. Finally, the mix activity $\textbf{G}^{(2)}$ is calculated by convolving $\textbf{U}^{(2)}$ with $\textbf{H}^{(1)}[t]$. After the model has run for a specified number of timesteps, the updated $\textbf{H}^{(2)}[t]$ is passed to the next layer.

\begin{figure}[t!]
\begin{center}
   \includegraphics[width=.98\linewidth]{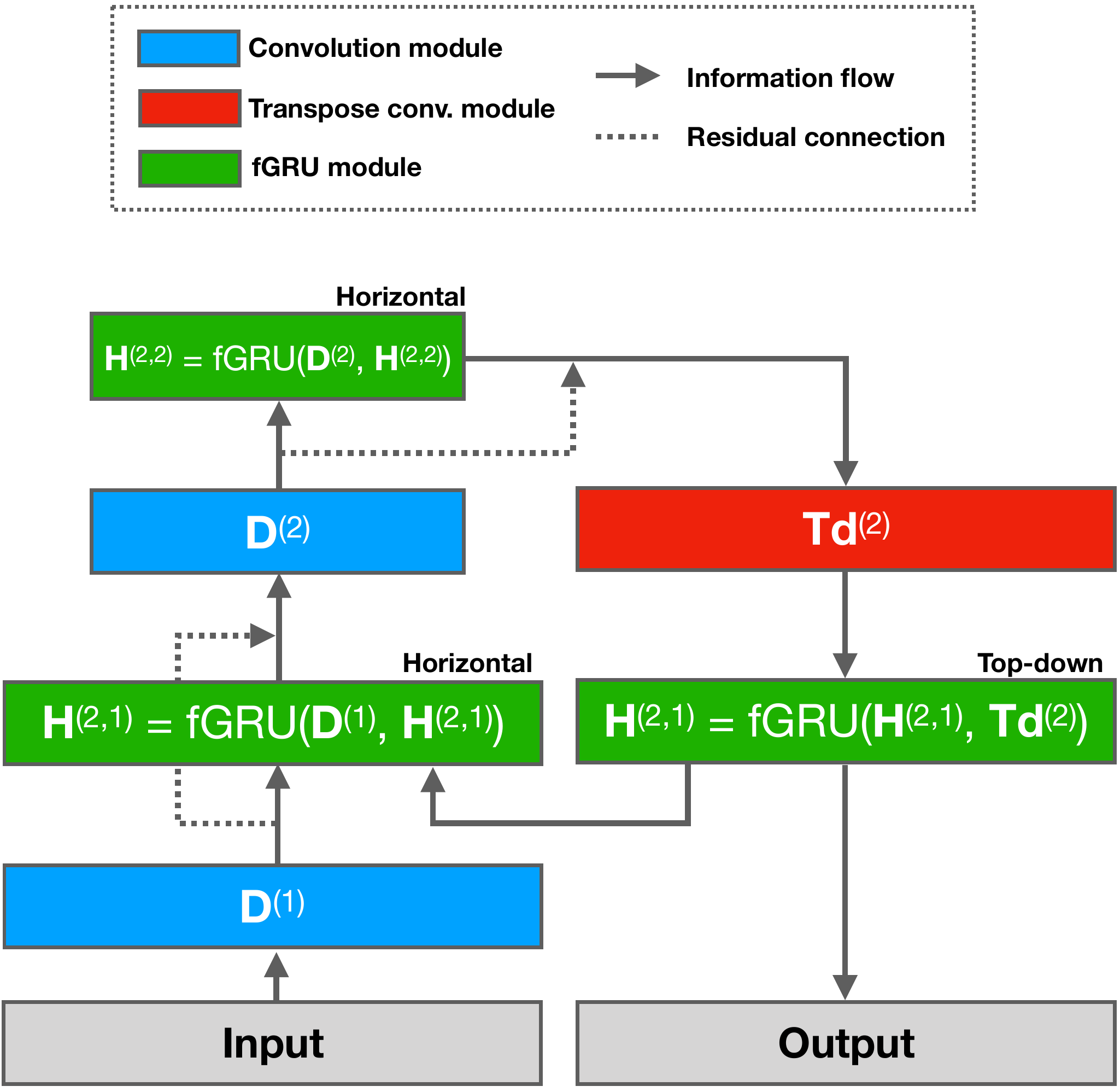}
\end{center}%\vspace{-4mm}
   \caption{A high-level diagram of one timestep of processing in a two-layer fGRU (fGRU-2). Information flows serially through each layer of the model on every timestep, implementing a bottom-up and then top-down pass of activity. Blue rectangles depict convolutional modules (D), each of which consists of a convolutional layer, batch-norm. Max pool subsampling is shown by a reduced width of the rectangle for the following layer. The fGRU-2 features a transpose convolutional module, shown with the red rectangle, that upsamples activity from the second fGRU to match the height and width of the first. The three fGRU modules, shown in green rectangles, control the low-level horizontal interactions, high-level horizontal interactions, and top-down feedback from high-to-low level fGRU modules (ordered from the input to the output). Residual connections are added to improve learning, and denoted with dotted lines.}
\label{fig:hgru_diagram}
\end{figure}

\paragraph{Proposed fGRU module}
A single-layer hGRU model efficiently solves a toy contour detection task by learning horizontal connections that implement a visual strategy of contour integration~\cite{Linsley2018-ls}. However, as we demonstrate in section~\ref{results}, this model is suboptimal on the more challenging problem of neuron segmentation. Indeed, cognitive neuroscience suggests that top-down feedback from higher-processing regions is a key ingredient for perceptual grouping that is missing from this hGRU~\cite{Roelfsema2016-vy, Parent1989-hr}. Higher-level feature information can regulate and constrain the horizontal connections between units at a lower processing stage to efficiently group task-relevant image features~\cite{Gilbert2013-hb}. A complementary perspective casts these low-level units as a recurrent ``scratch-pad'' for higher-level regions, which is directly read-out from to make decisions about the world after it is sufficiently processed by top-down feedback~\cite{Gilbert2007-hr,Slotnick2005-gd}. Motivated by these considerations, we introduce a generalized expression of the hGRU module, the feedback gated recurrent (fGRU), which utilizes the dynamics of the hGRU in Eq.~\ref{hGRU} to construct a model with both horizontal and top-down feedback connections.

The hGRU takes two inputs on every step of processing, \textbf{X} and $\textbf{H}^{(2)}[t-1]$, corresponding to a feedforward drive from a prior layer, and the persistent output activity of the hGRU. An fGRU module can implement top-down feedback by relaxing constraints on the origins of these activities.

\begin{align}\vspace{-1mm}
    \begin{split}
    \textbf{H}^{(2,l)}[t] &= \mathrm{fGRU}(\textbf{H}^{(2,l)}[t], \textbf{H}^{(2,l+1)}[t])\\
    \end{split}
\end{align}%\vspace{-5mm}

The function fGRU applies the hGRU dynamics in Eq.~\ref{hGRU} to two input activities (Fig.~\ref{fig:hgru_diagram}). The feedforward drive input \textbf{X} of an hGRU module is substituted with $\textbf{H}^{(2,l)}[t]$ which is the recently updated output of the fGRU module at layer $l$. Likewise, the output activity $\textbf{H}^{(2)}[t-1]$ of an hGRU module is replaced with $\textbf{H}^{(2,l+1)}[t]$ which denotes the output of another fGRU module at the subsequent layer $l+1$. An fGRU receiving these inputs will therefore learn to apply feedback from the high-level output (layer $l+1$) onto units in its low-level (layer $l$). On every timestep of processing, the gain activity selects low-level feature channels for processing, and the mix integrates the resulting activity with high-level horizontal activities. Because the fGRU is a generalization of the hGRU for feedback, it is important to note that a special case application of the fGRU can learn horizontal connections: $\textbf{H}^{(2,l)}[t] = \mathrm{fGRU}(\textbf{X}^{l}, \textbf{H}^{(2,l)}[t])$.

%The equations here can model either horizontal or top-down connections. Take for example the ``Top-down'' operation presented in Fig~\ref{fig:arch_diagram}: the

\begin{figure*}[t!]
\begin{center}
   \includegraphics[width=.90\linewidth]{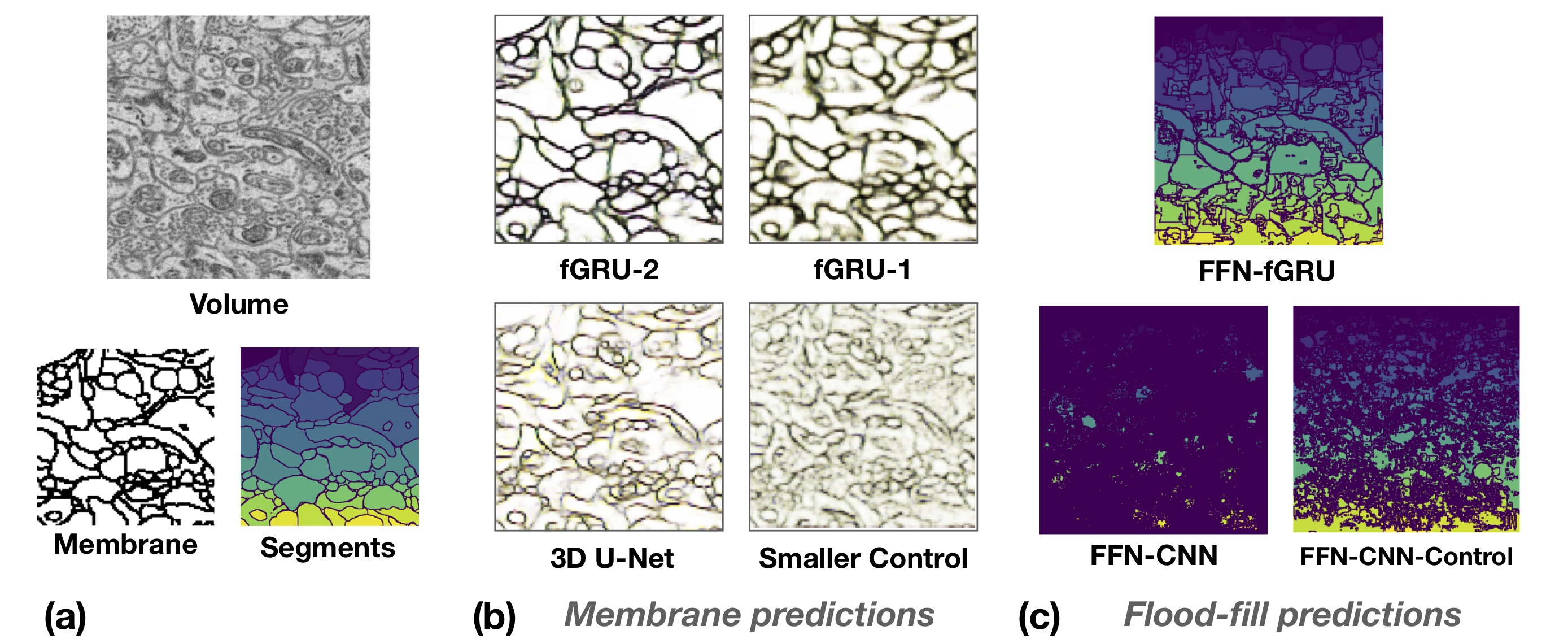}
\end{center}%\vspace{-4mm}
   \caption{fGRU networks outperform a state-of-the-art 3D U-Net~\cite{Lee2017-ye} for membrane detection on the STAR challenge. (a) An exemplar image, membrane labeling, and neurite segmentation from the STAR test volume. (b) Membrane predictions of fGRUs, a 3D U-Net~\cite{Lee2017-ye}, and a non-recurrent CNN with a similar number of parameters as the fGRU (Smaller Control). Model predictions are plotted as unthresholded nearest-neighbor affinity estimates in the X/Y/Z directions; higher-precision estimates result in high-contrast depictions of the membrane predictions. (c) Segmentation predictions from models based on flood-filling networks~\cite{Januszewski2018-rl}. These are the reference FFN-CNN, a FFN-fGRU (an fGRU replacing its convolutional layers), and a non-recurrent FFN-CNN with a similar number of parameters as the FFN-fGRU (FFN-CNN-Control).}
\label{fig:results1}
\end{figure*}

\section{The STAR challenge}\label{results}
% \paragraph{Challenge}
The STAR challenge introduces a labeled image stack of mouse retina from~\cite{Briggman2011-aw}, which is $384\times384\times384$ voxels. Annotations are kept private for the challenge, and performance is evaluated on the private server at \url{STAR-challenge.github.io}. The STAR training set consists of labeled volumes of fruit fly nervous system and mouse cortex. These datasets come from the MICCAI circuit reconstruction of electron microscopy images (CREMI), Segmentation of neurites in EM images (SNEMI3D;~\cite{Kasthuri2015-fh}), and FIB-25 (see Table~\ref{tab:datasets}).

\paragraph{Training procedure}
We trained two classes of models on the STAR challenge, which represent the leading approaches for automated connectome reconstruction. One approach is to predict the locations of cell membranes in a volume, and use postprocessing routines (such as watershed and GALA~\cite{Nunez-Iglesias2013-eo}) to derive segmentations. The other approach is to train a model to segment one neuron in a volume at a time~\cite{Januszewski2018-rl}. The state of the art models of each approach are the 3D U-Net~\cite{Lee2017-ye} and flood-filling networks (FFN-CNN~\cite{Januszewski2018-rl}), respectively. Our implementations of these models are consistent with the published versions (see SI for details and validations). We also developed fGRU models and control models for each approach that were trained using identical training routines, which are discussed below. All models were trained using Tensorflow, and Adam~\cite{Kingma2014-ct}.

\paragraph{Membrane prediction models}
Membrane prediction models were trained according to the routine described in~\cite{Lee2017-ye}. Models were optimized for predicting the affinities of voxel pairs at a variety of distances, and a wide array of data augmentations were applied to the input volumes (only nearest-neighbor affinities were used for predictions; see SI for details). We focus on three membrane prediction models in addition to this reference 3D U-Net. First, we consider two models that use the fGRU module as an elementary component: (1) a one-layer fGRU model (fGRU-1) and (2) a two-layer fGRU (fGRU-2). We also consider a third model, a control CNN (FF-control) for understanding how the number of trainable parameters \textit{per se} influences membrane prediction on the STAR challenge. Each model was trained for at least 100K steps using the learning rate schedule of~\cite{Lee2017-ye}\footnote{Another 3D U-Net was trained with L2 weight decay, but it was less effective on the STAR challenge than the reference model (mAP = 0.35).}. Because the segmentation step in this approach is handled by hand-tuned post-processing routines, we only report membrane prediction performance here (but see Fig. \ref{fig:seung_validation} for segmentation performance of the reference 3D U-Net). 

The fGRU-2 consists of two sets of convolutional layers followed by fGRU modules, with an intervening max pool subsampling using a $4\times4\times1$ kernel (corresponding to height, width, and depth) and $4\times4$ stride (Fig.~\ref{fig:arch_diagram}). The convolutional layers both have $5\times5\times1$ kernels with 18 features. The kernel in the first fGRU module is $9\times9\times1$, the kernel in the second one is $5\times5\times3$, and both have 18 features. In other words, these are kernels which learn horizontal connections in each layer, and only the second layer's kernel learns these connections across three spatial dimensions. The fGRU modules also consist of separate sets of $1\times1\times1$ kernels $\textbf{U}^{(1)}, \textbf{U}^{(2)}$ with 18 features for calculating the gain and the mix in each fGRU module. A transpose convolution kernel with a $4\times4\times1$ kernel upsamples the second fGRU's persistent activity to the height and width of its first fGRU. An additional fGRU module learns to implement top-down feedback from the second fGRU onto the first with a $1\times1\times1$ kernel, resulting in the operation: $\textbf{H}^{(2,1)}[t] = \mathrm{fGRU}(\textbf{H}^{(2,1)}[t], \textbf{H}^{(2,2)}[t])$. The fGRU-1 has a convolutional layer with $5\times5\times1$ kernels and 18 features, along with an fGRU for learning horizontal connections with a $9\times9\times3$ kernel. Both models are run for 8 timesteps, after which the persistent activity of their first fGRU ($\textbf{H}^{(2,1)}$) is passed through batch-norm and a final convolutional classifier with a $1\times1\times1$ kernel to readout membrane predictions at every voxel.

Finally, as with the hGRU~\cite{Linsley2018-ls}, the fGRU module weight kernel $\textbf{W}$ is constrained to have symmetric weights between channels. This reduces the number of learnable parameters by nearly half \vs a typical convolutional kernel, and means that the weight $W_{x_0 + \Delta x, y_0 + \Delta y, k_1, k_2}$ is equal to the weight $W_{x_0 + \Delta x, y_0 + \Delta y, k_2, k_1}$, in which $x_0$ and $y_0$ denote the center of the kernel.
% \vspace{-2mm}

%%%%%%%%%%%%%%%%%%%%%%
% Performance table
%%%%%%%%%%%%%%%%%%%%%%
\begin{table}[t!]
\centering
\begin{tabular}{|l|c|c|}
% \hline & mAP & ARAND & Parameters\\ \hline
% FF-Control & 0.44 & X & 60K\\
% 3D U-Net~\cite{Lee2017-ye} & 0.51 & XXX & 1.4M\\
% fGRU-1 & 0.60 & I & 45K\\
% fGRU-2 & \textbf{0.72} & \textbf{I} & 50K\\\hline
% FFN-CNN~\cite{Januszewski2018-rl} & - & XXX & 630K\\
% FFN-fGRU & - & \textbf{I} & 100K\\\hline
\hline & mAP & Parameters\\ \hline
FF-control & 0.44 & 60K\\
3D U-Net~\cite{Lee2017-ye} & 0.51 & 1.4M\\
fGRU-1 & 0.60 & 45K\\
fGRU-2 & \textbf{0.72} & 50K\\\hline
\end{tabular}
\linebreak\linebreak%\vspace{-4mm}
\caption{Membrane prediction performance on the STAR challenge. Membrane predictions are evaluated by mean Average Precision (mAP; higher is better). For reference, a 3D-Unet~\cite{Lee2017-ye} trained and tested on distinct sections of the STAR test volume scores 0.78 mAP.}
\label{tab:membrane_pred}%\vspace{-5mm}
\end{table}
%%%%%%%%%%%%%%%%%%%%%%
%%%%%%%%%%%%%%%%%%%%%%

\paragraph{Membrane prediction performance}

We first performed a validation analysis to measure each model's ability to predict membranes in the STAR test volume. Models were trained to predict voxel affinities on 80\% of the STAR test volume images with single volume batches of size $160 \times 160 \times 18$ (X$\times$Y$\times$Z dimensions) and tested on the remaining 20\%. The fGRU-2 was on par with the 3D U-Net in predicting these membranes, and both of these models outpaced the others (see Table~\ref{tab:berson_performance_si} in SI for results). Next, we used the same routine to train each model on samples from 80\% of the STAR training dataset; 20\% of it was held out for validation, which was used to select model weights for subsequent testing (Fig.~\ref{fig:results1}a). When evaluated on the STAR test dataset, the 3D U-Net's performance decreased by 0.27 points of mean average precision (Table~\ref{tab:membrane_pred}). This precipitous drop is evident in the model's predictions: it confuses organelles in the SEM images with cell membranes, and fails to accurately group together continuous sections of cell membrane (Table~\ref{tab:membrane_pred} and Fig.~\ref{fig:results1}b). By contrast, the fGRU-1 fairs significantly better and more consistently defines cell membranes.

\begin{figure}[t]
\begin{center}
   \includegraphics[width=1\linewidth]{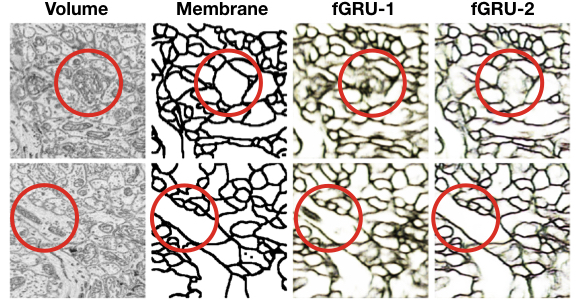}
\end{center}%\vspace{-4mm}
   \caption{Top-down feedback gates the horizontal excitation and inhibition of low-level units. Exemplar images from the STAR test volume along with membrane labels and predictions from a 1- and 2-layer fGRU. Red circles demark regions in which the fGRU-1 shows uncontrolled excitation that spills into cell membranes. Top-down connections in the fGRU-2 correct this problem by using higher-level feature information to modulate low-level horizontal connections.}
\label{fig:feedback}
\end{figure}

The fGRU-2 further improves on the performance of the fGRU-1 (Table~\ref{tab:membrane_pred}). Visual inspection of differences in predictions between these two models reveals that the fGRU-2 is more effective at suppressing irrelevant visual features within cells (Fig.~\ref{fig:feedback}). When learning horizontal connections, the fGRU-1 links together units in its feedforward drive that encode collinear features across long spatial distances, such as units that overlap an elongated membrane. A downside to this strategy is that it is unconstrained and can lead to excitation of task-irrelevant features. Top-down feedback in the fGRU-2 helps correct errors like ``runaway'' excitation by introducing high-level contextual information into the persistent activity of a low-level fGRU (red circles in Fig.~\ref{fig:feedback} highlight examples of such inhibition).

One possible explanation for fGRU-based models outperforming the 3D U-Net on this challenge is that they contain only a fraction of the number of parameters (Table~\ref{tab:membrane_pred}, Parameters). In other words, it is possible that parameter efficiency \textit{per se} might predict performance on the STAR. To answer this question, we trained an additional feedforward model that replaced the fGRU modules in the fGRU-2 with 3D convolutional layers (kernel size: $7\times7\times3$), and removed both subsampling and upsampling, resulting in a similar number of parameters as the fGRU-2 (see SI for details\footnote{We trained another control with a similar number of parameters in a 3D U-Net architecture, but it was less effective than this model.}). This model was neither effective in the validation analysis of training and testing on the STAR test set (Table S1 in SI), nor did it perform well in the STAR challenge (Table~\ref{tab:membrane_pred} and Fig.~\ref{fig:results1}b).

\begin{figure}[t!]
\begin{center}
   \includegraphics[width=1\linewidth]{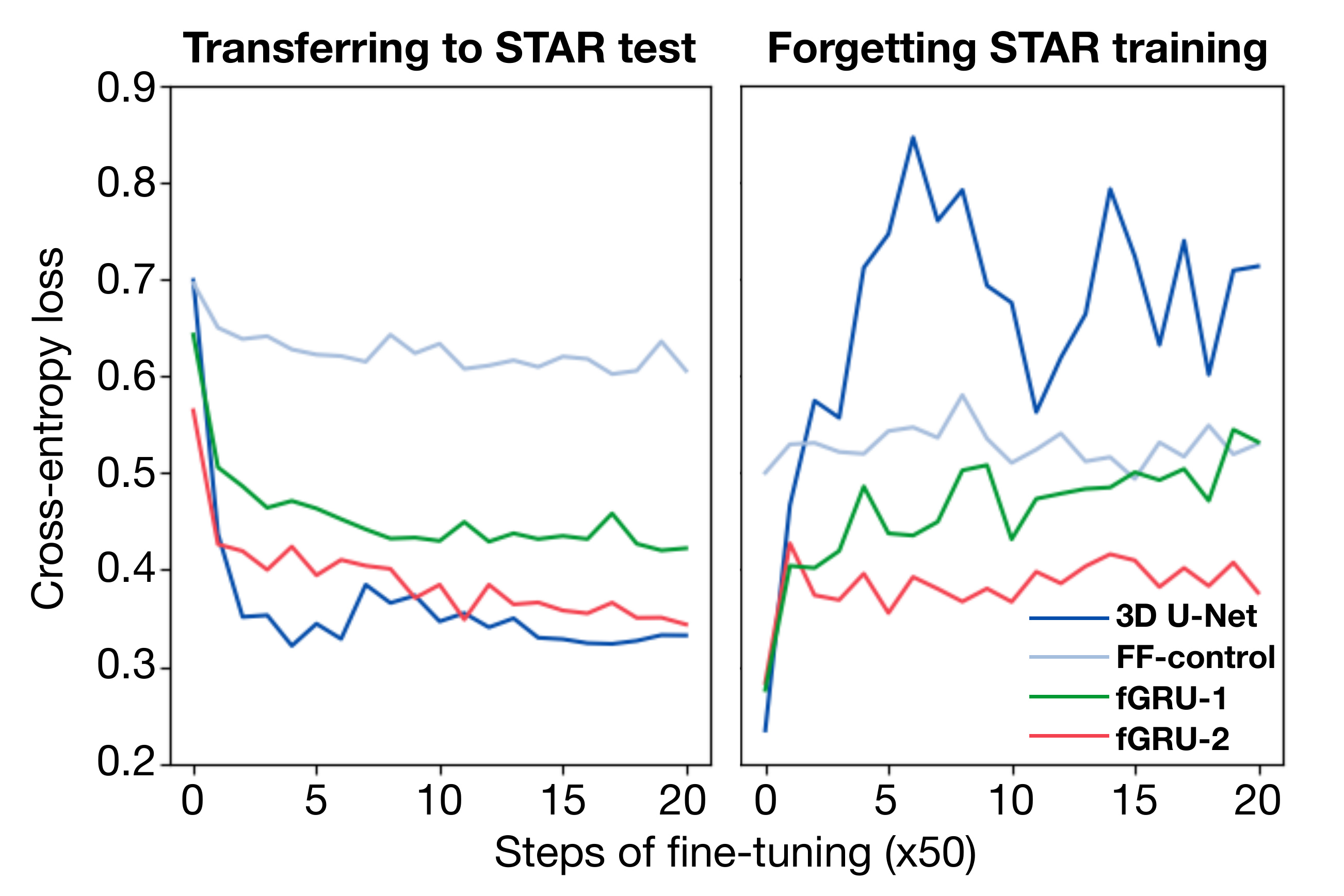}
\end{center}%\vspace{-4mm}
   \caption{Forgetting correlates with model performance on the STAR challenge. Models are trained for the STAR challenge before their weights are ``fine-tuned'' to a cross-validation split (80/20) of the Briggman test volume. During fine-tuning, mean per-voxel crossentropy loss is recorded on the held-out split of the Briggman volume, as well as on a similar sized split from a volume in the STAR training dataset. All models become more effective at predicting membranes in the STAR test dataset from this transfer learning. fGRU models show less ``forgetting'' on a volume in the STAR training set (FIB-25), suggesting that they are less vulnerable to overfitting than the other models.}
\label{fig:membrane_forgetting}
\end{figure}

Is there a metric that predicts model performance? After training these models on the STAR challenge, we investigated their ability to learn the STAR test dataset via fine-tuning (Fig.~\ref{fig:membrane_forgetting}, left panel). At the same time, we recorded model performance in predicting membranes on the STAR training dataset (Fig.~\ref{fig:membrane_forgetting}, right panel). These performance curves indicate that the 3D U-Net is very amenable to fine-tuning, and rapidly learns to precisely predict membranes in the STAR test dataset. However, during fine-tuning, it completely ``forgets'' the original dataset. The fGRU-1 shows a similar pattern of results, but with less effective transfer and less severe forgetting. The fGRU-2, on the other hand, transfers nearly as quickly as the 3D U-Net while also maintaining information about the STAR training dataset. Together, these results suggest that a model's tendency towards ``catastrophic forgetting''~\cite{Masse2018-cj} is a useful metric for measuring overfitting, and in turn, predicting its ability to generalize to the STAR test set.

\paragraph{Flood-filling models} Flood-filling models were trained according to the routine described in~\cite{Januszewski2018-rl} with slight variations to improve the speed of convergence (see SI for details). We trained three different versions of this model: (i) the publicly available reference version from~\cite{Januszewski2018-rl} (FFN-CNN); (ii) a version with an fGRU in place of its convolutional network; (iii) a ``shallower'' variant of the FFN-CNN to study the effect of reduced feature depth and overall network size (FFN-CNN-Control). See SI for details on model architectures and training routines.

%%%%%%%%%%%%%%%%%%%%%%
% Performance table
%%%%%%%%%%%%%%%%%%%%%%
\begin{table}[t!]
\centering
\begin{tabular}{|l|c|c|c|}
% \hline & mAP & ARAND & Parameters\\ \hline
% FF-Control & 0.44 & X & 60K\\
% 3D U-Net~\cite{Lee2017-ye} & 0.51 & XXX & 1.4M\\
% fGRU-1 & 0.60 & I & 45K\\
% fGRU-2 & \textbf{0.72} & \textbf{I} & 50K\\\hline
% FFN-CNN~\cite{Januszewski2018-rl} & - & XXX & 630K\\
% FFN-fGRU & - & \textbf{I} & 100K\\\hline
\hline & Precision & Recall & Parameters\\ \hline
FFN-CNN~\cite{Januszewski2018-rl} & 0.772 & 0.057 & 630K\\
FFN-CNN-Control & 0.821 & 0.063 & 196K\\
FFN-fGRU & \textbf{0.917} & \textbf{0.414} & 126K\\\hline
\end{tabular}
\linebreak\linebreak \vspace{-4mm}
\caption{Segmentation performance on the STAR challenge, evaluated by ``patchwise'' precision and recall (higher is better). These metrics are computed as the mean precision and recall of predicted cell masks over 500 iterations of network processing in randomly sampled fields of vision (hence, ``patchwise''). The final segmentation produced by flood-filling networks depends on hand-tuned hyperparameters. Patchwise precision-recall better reflects the quality of the models' cell segmentations because it does not depend on such hand-tuning.}
\label{tab:ffnet_pred}
\end{table}
%\vspace{-5mm}
%%%%%%%%%%%%%%%%%%%%%%
%%%%%%%%%%%%%%%%%%%%%%

\paragraph{Flood-filling performance} These models were first validated on their ability to segment neurons in the STAR test set when trained on a held-out section of the same volume (see Table~\ref{tab:ffn_performance_si} in SI for results): FFN-CNN and FFN-fGRU performed similarly, and both outperformed the FFN-CNN-Control. We next trained these models on the STAR challenge, and then tested them on the STAR test volume (Table~\ref{tab:ffnet_pred}). The FFN-CNN did not generalize to this volume. Simply using a smaller and shallower network did not rescue performance either: the FFN-CNN-Control was also poor at segmenting this test volume. The FFN-fGRU, on the other hand, produced sensible, albeit oversegmented, predictions (Fig.~\ref{fig:results1}). 

\section{Discussion}
We have presented STAR, a novel connectomics challenge, aimed at testing the ability of architectures for neuron reconstruction to generalize to novel brain tissues beyond those used for training. We have shown that state-of-the-art architectures perform well when trained and tested on different subsets of the same volume but generalize poorly to different volumes. That convolutional neural networks (CNNs) exhibit a limited ability to generalize beyond training data is known~\cite{Azulay2018-lp,Zhang2016-or,Jo2017-ay,Geirhos2017-tm,Wang2017-sl,Saleh2016-qa,Rosenfeld2018-jn} but often overlooked. % The present study shows that CNNs do not generalize well to qualitatively small differences in cell morphology across brain tissues and slight image perturbations that arise because of variations in the imaging equipment used, even when these models are trained with data augmentations that might be expected to build tolerance to such differences. 
Our results are consistent with a recent study that demonstrated that CNNs trained for image categorization on CIFAR do not generalize well to images from a novel test set even though the training and test sets had near identical statistics~\cite{Recht2018-ix}. Similarly, it has been shown that while CNNs can be trained to handle specific kinds of noise, they are unable to generalize to unseen (albeit similar) noise conditions~\cite{Geirhos2017-tm}.

We have further described a novel deep learning module, the feedback gated recurrent unit (fGRU), which leverages recurrent interactions -- via both short-range horizontal connections within a convolution layer and long-range top-down connections between layers. The architecture was shown to generalize better than all tested baselines. Our work further adds to a growing body of literature which suggests that, consistent with the anatomy and physiology of the visual cortex, feedback connections may be critical to visual recognition tasks. Previous work has shown how feedback may provide prior information in predictive coding models~\cite{Han2018-wu, Lotter2016-qr, OReilly2017-ee} and help solve contour tracing, incremental grouping, object recognition performance~\cite{George2017-ae, Sabour2018-rg, Spoerer2017-ee, Tang2018-vf, Nayebi2018-dc}, and other visual recognition tasks that require learning long-range statistical dependencies~\cite{Brosch2015-sr, Linsley2018-ls}. The present study also demonstrates the limitations of the horizontal interactions implemented by one of these models, the hGRU~\cite{Linsley2018-ls}, which are strained by the visual complexity of neuron segmentation in SEM images. The fGRU corrects this deficiency by controlling horizontal connections at a lower layer with top-down feedback connections from a higher one, which we find helps to suppress some of the pathological dynamics that emerge in a single hGRU layer. One possible functional explanation for this is that the fGRU disambiguates multiple interpretations of the visual image at a lower layer by introducing contextual information from a higher layer. 

In summary, this work advocates for shifting the emphasis of computer vision in connectomics from fitting volume-specific datasets to generalizing across volumes. It is our hope that the STAR challenge and fGRU model framework constitute building blocks towards this goal.

\subsection{Acknowledgements}
We are grateful to the authors of \cite{Januszewski2018-rl} who have helped us implement their model and Kalpit Thakkar who created the challenge website. This research was supported by the Carney Institute for Brain Science through an innovation award granted to TS and DB. Additional support was provide by the Brown University Center for Computation and Visualization (CCV), as well as Meridian IT and IBM.

%-------------------------------------------------------------------------

\clearpage
% \end{document}
% \begin{document}

%% Prepare SI figures and counters
\setcounter{figure}{0}

\makeatletter 
\renewcommand{\thefigure}{S\@arabic\c@figure}
\makeatother

\setcounter{table}{0}

\makeatletter 
\renewcommand{\thetable}{S\@arabic\c@table}
\makeatother

% \vskip .375in
\twocolumn[\begin{center}
{\Large \bf Supplementary Information: Robust neural circuit reconstruction from serial electron microscopy with convolutional recurrent networks \par}
% additional two empty lines at the end of the title
\vspace*{24pt}
\large
\lineskip .5em
\end{center}]

\section{Membrane prediction models}
\paragraph{Training and testing}\label{sec:membrane_training}
Our reference model for membrane prediction is the 3D U-Net of~\cite{Lee2017-ye}. We followed their published routine for training all membrane prediction models. Key to this is their use of a large set of random data augmentations applied to SEM image volumes, which simulate common noise and errors in SEM imaging. These are (i) misalignment between consecutive $z$-locations in each input image volume. (ii) Partial- or fully-missing sections of the input image volumes. (iii) Blurring of portions of the image volume. Augmentations that simulated these types of noise, as well as random flips over the $xyz$-plane, rotations by $90^{\circ}$, brightness and contrast perturbations, were applied to volumes following the settings of~\cite{Lee2017-ye}. Models were trained to convergence, for at least 100K iterations, using Adam~\cite{Kingma2014-ct} and the learning rate schedule of~\cite{Lee2017-ye}, in which the optimizer step-size was halved when validation loss stopped decreasing (up to four times). Models were trained with single-SEM volume batches of $160\times160\times18$ (X/Y/Z), normalized to $[0, 1]$. All models were trained to predict nearest-neighbor voxel affinities, as well as 3 other mid- to long-range voxel distances, as in~\cite{Lee2017-ye}. Only the nearest neighbors were used at test time. 

\begin{figure}
\begin{center}
   \includegraphics[width=1.\linewidth]{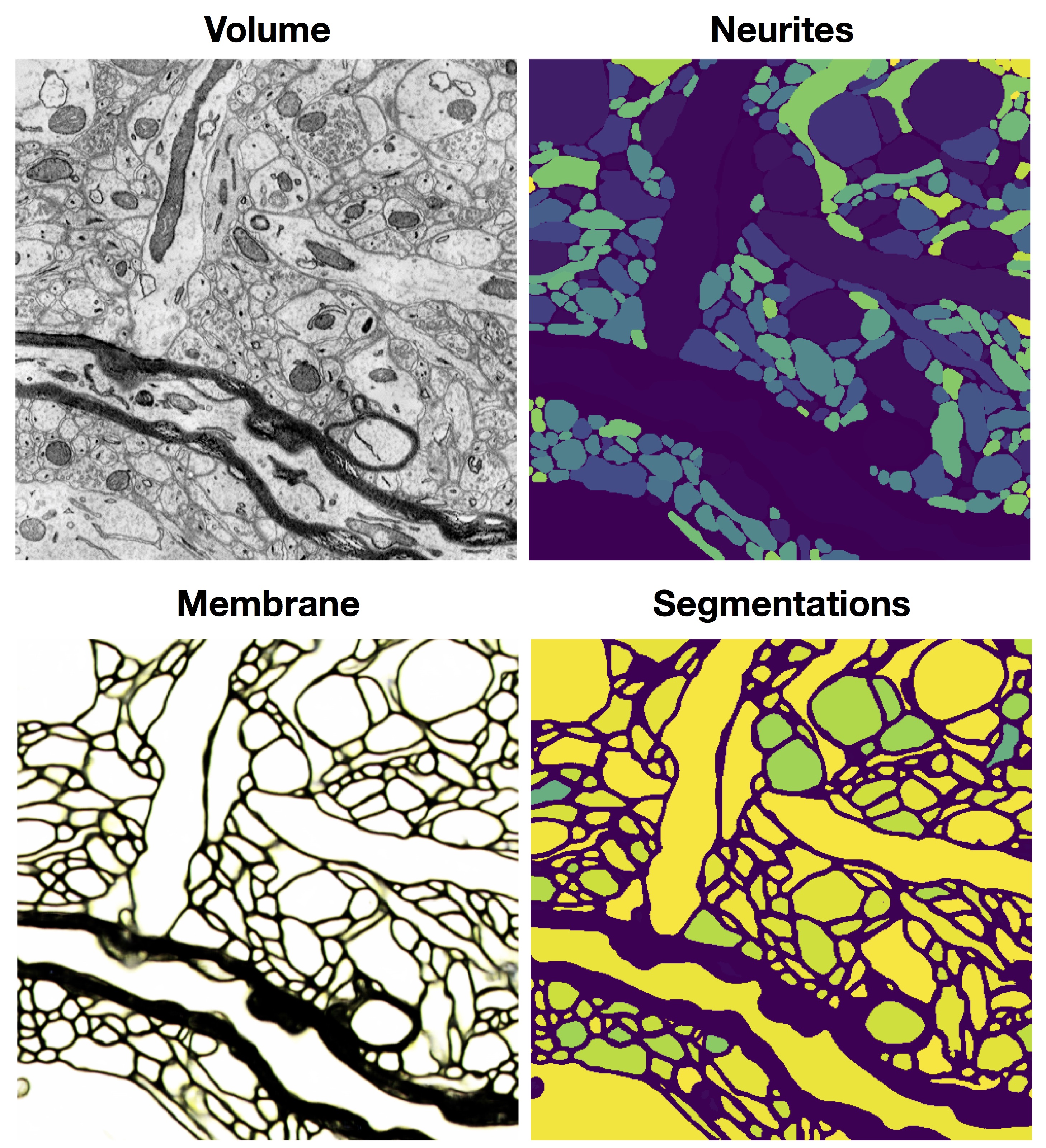}
\end{center}%\vspace{-4mm}
   \caption{We trained the reference 3D U-Net from~\cite{Lee2017-ye} on the SNEMI3D dataset to validate the implementation. Segmentations here are derived by watershedding and agglomeration with GALA~\cite{Nunez-Iglesias2013-eo}, resulting in ``superhuman'' ARAND (evaluated according to the SNEMI3D standard; lower is better) of 0.04, which is below the reported human-performance threshold of 0.06 and on par with the published result (see Table 1 in~\cite{Lee2017-ye}, mean affinity agglomeration).}
\label{fig:seung_validation}
\end{figure}

\begin{figure*}[t]
\begin{center}
   \includegraphics[width=.6\linewidth]{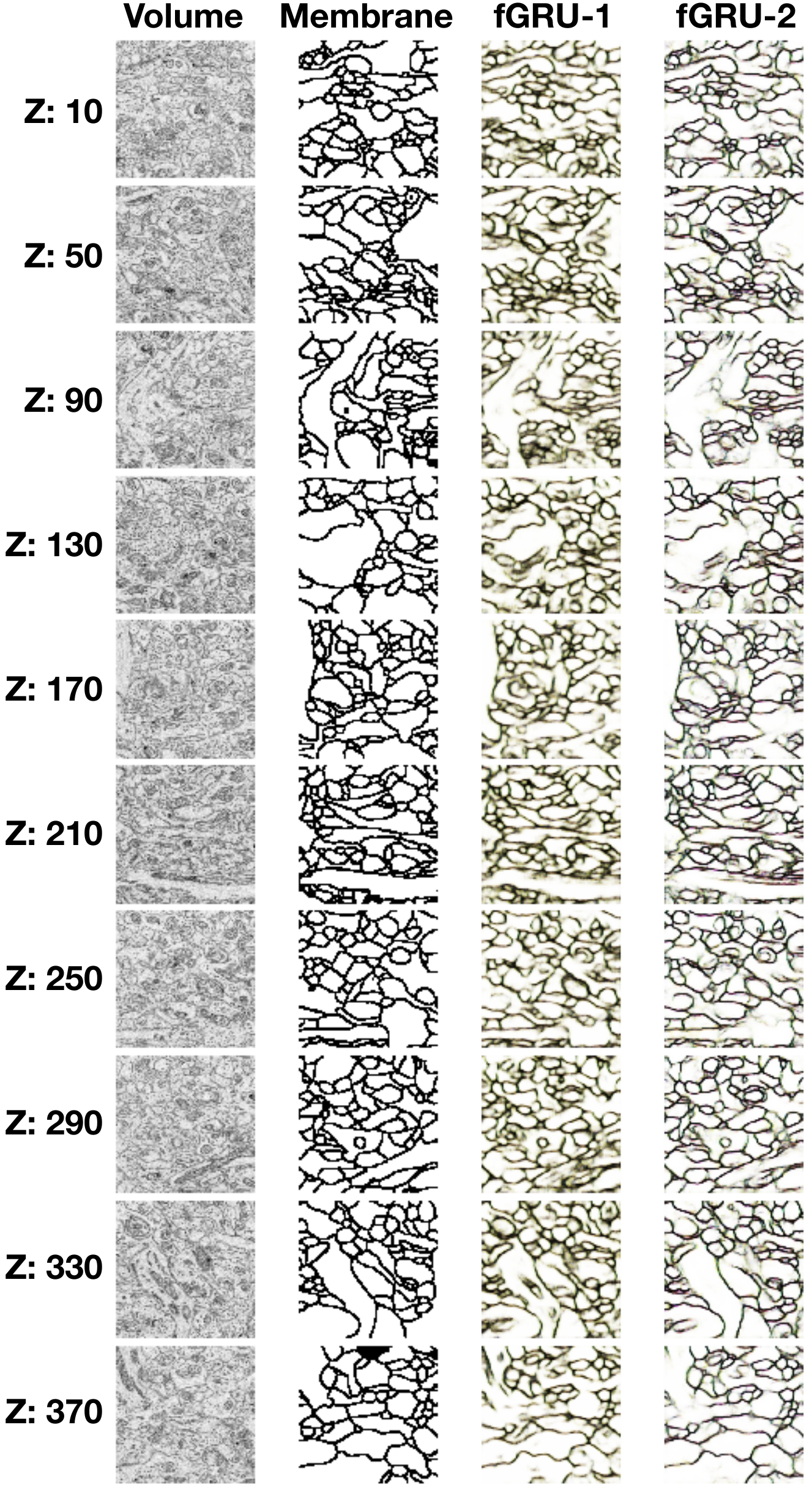}
\end{center}%\vspace{-4mm}
   \caption{An extended comparison of membrane prediction performance for the one- and two-layer fGRU models (fGRU-1 and fGRU-2) on images from the STAR test dataset. Note the increased sparsity of the fGRU-2 predictions \vs the fGRU-1. The higher contrast of the fGRU-2 predictions indictates greater confidence than the fGRU-1.}
\label{fig:membrane_forgetting_si}
\end{figure*}

We performed 3D neuron reconstructions on membrane predictions using a pipeline consisting of watershed to derive supervoxels and GALA~\cite{Nunez-Iglesias2013-eo} for agglomerating these into precise segmentations. We used this approach to validate our implementation of the 3D U-Net~\cite{Lee2017-ye} (Fig.~\ref{fig:seung_validation}).

% , and to establish the fGRU-2's state-of-the-art ARAND score on the STAR challenge of 0.XX (the GALA model used in this case was trained on segmentations from the SNEMI3D challenge).

We validated membrane prediction models on the STAR challenge test set. Models were trained on the first 80\% of the volume's $z$-axis images, and tested on the remainder. Results are shown as mean average precision (mAP; scores calculated per $z$-axis image and then averaged together) in Table~\ref{tab:berson_performance_si}.

\paragraph{fGRU membrane prediction}

After training fGRU-based models on the STAR challenge, we investigated how the top-down feedback of the fGRU-2 influenced its predictions. Exemplar slices from the STAR test dataset are depicted in Fig.~\ref{fig:membrane_forgetting_si}, along with the ground-truth membrane labels, and per-pixel membrane predictions from a 1- and 2-layer fGRU. The most apparent differences between the two models is that the top-down feedback of the 2-layer fGRU has stronger control over the incorrect extension of membrane predictions into cell parts. These results, which depict nearest-neighbor affinity predictions, also reveal that the fGRU-2 was more confident in its membrane predictions than the fGRU-1.

\section{Flood-filling models}

%%%%%%%%%%%%%%%%%%%%%%
% Performance table berson
%%%%%%%%%%%%%%%%%%%%%%
\begin{table}[t]
\centering
\begin{tabular}{|l|c|c|}
\hline & mAP & Parameters\\ \hline
FF-Control & 0.63 & 60K\\
fGRU-1 & 0.74 & 45K\\
fGRU-2 & 0.77 & 50K\\
3D U-Net~\cite{Lee2017-ye} & \textbf{0.78} & 1.4M\\\hline
\end{tabular}
\linebreak\linebreak%\vspace{-4mm}
\caption{Reference performance for models after training and testing on distinct image splits (80\%/20\%) of the STAR test volume. Membrane predictions are evaluated by mean Average Precision (mAP; higher is better).}
\label{tab:berson_performance_si}%\vspace{-5mm}
\end{table}
%%%%%%%%%%%%%%%%%%
% End performance table
%%%%%%%%%%%%%%%%%%

\begin{figure}[t!]
\begin{center}
   \includegraphics[width=.98\linewidth]{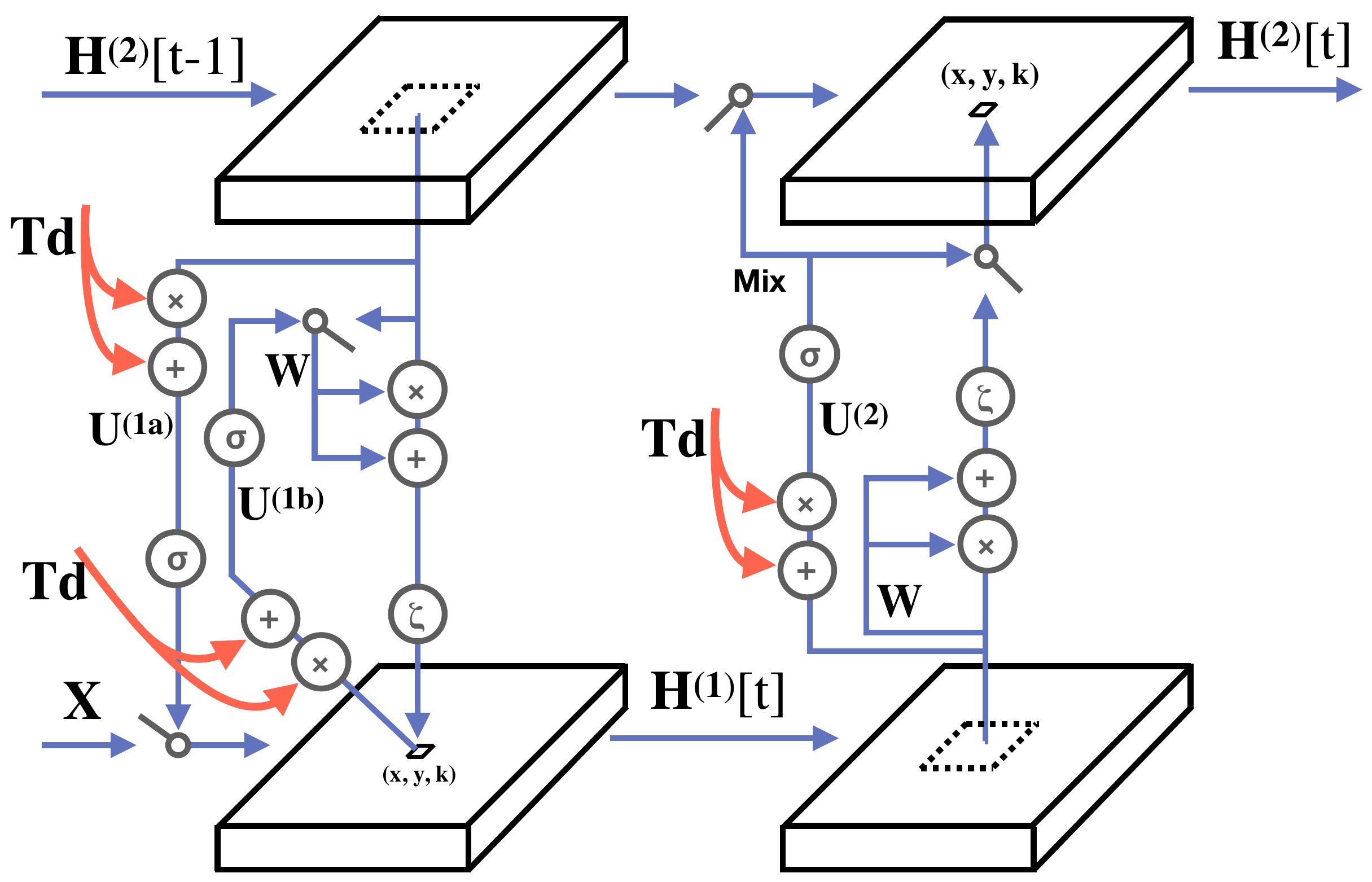}
\end{center}%\vspace{-4mm}
   \caption{A diagram of the fGRU* module used in the FFN-fGRU. This module uses top-down feedback ($\textbf{Td}$) to compute its gains, $\textbf{G}^{(1a)}, \textbf{G}^{(1b)},$ and $\textbf{G}^{(2)}$. In this module, $\textbf{Td}$ helps to gate horizontal interactions implemented by the kernel $\textbf{W}$. In addition, we found that introducing an extra gate that controls the flow of $\textbf{X}$ improved performance in neuron segmentation. This allows the network to selectively modulate its feedforward drive while calculating updates to its hidden state $\textbf{H}^{(2)}$.}
\label{fig:fgru_st_lowlevel}
\end{figure}

\paragraph{The fGRU* module}\label{fgru*} We discovered that several modifications to the original fGRU module led to significantly better generalization performance when deriving neuron segmentations with the flood-filling network approach (Fig.~\ref{fig:fgru_st_lowlevel}). These modifications result in a module called fGRU*, which makes the following changes to the fGRU: (i) it processes horizontal (\ie persistent activity at layer $l$) and top-down interactions (\ie persistent activity at layer $l + 1$) within a single module. By constrast, the fGRU discussed in the main text only processes one of these interactions at a time. (ii) Inspired by~\cite{Gilbert2013-hb}, the module uses top-down feedback to compute its gain and mix activities. (iii) The module includes a separate gain control activity to selectively modulate the flow of the feedforward drive $\textbf{X}$ into its hidden state $\textbf{H}^{(2)}$. This additional gain $\textbf{G}^{(1b)}$ joins the existing gain $\textbf{G}^{(1a)}$ (which is applied to the module's persistent activity). We describe the fGRU* with the following equations:
% This makes the module take three inputs (feedforward drive $\textbf{X}$, hidden state $\textbf{H}^{(2)}$ and feedback $\textbf{Fb}$) instead of two; (
\vspace{-1mm}
\begin{align}
\begin{split}
    \textbf{G}^{(1a)}[t] =& \sigma(\mathrm{BN}(\textbf{U}^{(1a)} * \\ & (\kappa_{1}(\textbf{H}^{(2)}[t-1]\odot\textbf{Td}) + \kappa_{2}\textbf{H}^{(2)}[t-1] + \kappa_{3}\textbf{Td})))\\
    \tilde{\textbf{X}} =&\textbf{X}\odot\textbf{G}^{(1a)}[t]\\
    \textbf{G}^{(1b)}[t] =& \sigma(\mathrm{BN}(\textbf{U}^{(1b)} * (\eta_{1}(\tilde{\textbf{X}}\odot\textbf{Td}) + \eta_{2}\tilde{\textbf{X}} + \eta_{3}\textbf{Td})))\\
    \textbf{C}^{(1)}[t] =& \mathrm{BN}(\textbf{W} * (\textbf{G}^{(1b)}[t] \odot \textbf{H}^{(2)}[t-1]))\\
    \textbf{G}^{(2)}[t] =& \sigma(\mathrm{BN}(\textbf{U}^{(2)} * \\ & ( \lambda_{1}(\textbf{H}^{(1)}[t]\odot\textbf{Td}) + \lambda_{2}\textbf{H}^{(1)}[t] + \lambda_{3}\textbf{Td}))))\\
    \textbf{C}^{(2)}[t] =& \mathrm{BN}(\textbf{W} * \textbf{H}^{(1)}[t])
\end{split}
\end{align}

The activities $\textbf{H}^{(1)}[t], \tilde{\textbf{H}}^{(2)}[t]$ and $\textbf{H}^{(2)}[t]$ are calculated identically to the fGRU described in the main text. In the fGRU*, the gains $\textbf{G}^{(1a)}$ and $\textbf{G}^{(1b)}$, and the mix $\textbf{G}^{(2)}$ are computed via 1$\times$1$\times$1 convolution using kernels $\textbf{U}^{(1a)}$, $\textbf{U}^{(1b)}$ and  $\textbf{U}^{(2)}$, respectively. These activities are computed with a combination of top-down feedback (\ie, persistent activity from layer $l+1$; $\textbf{Td}$) with input or horizontal activity (\ie, $\textbf{X}$ or persistent activity from layer $l$; $\textbf{H}^{(2)}$). The calculation of these activities also includes additional degrees of freedom \vs the fGRU of the main text. A learned scalar is included in the calculation of each to balance between linear (additive) and quadratic (multiplicative) combinations of horizontal and top-down activity: (i) linear and quadratic contributions to calculation of $\textbf{G}^{(1a)}$ are modulated by $\kappa$; (ii) for $\textbf{G}^{(1b)}$, $\eta$ has the same function; and (iii) for the mix, $\lambda$ performs this role. 

\begin{figure}[t!]
\begin{center}
   \includegraphics[width=.98\linewidth]{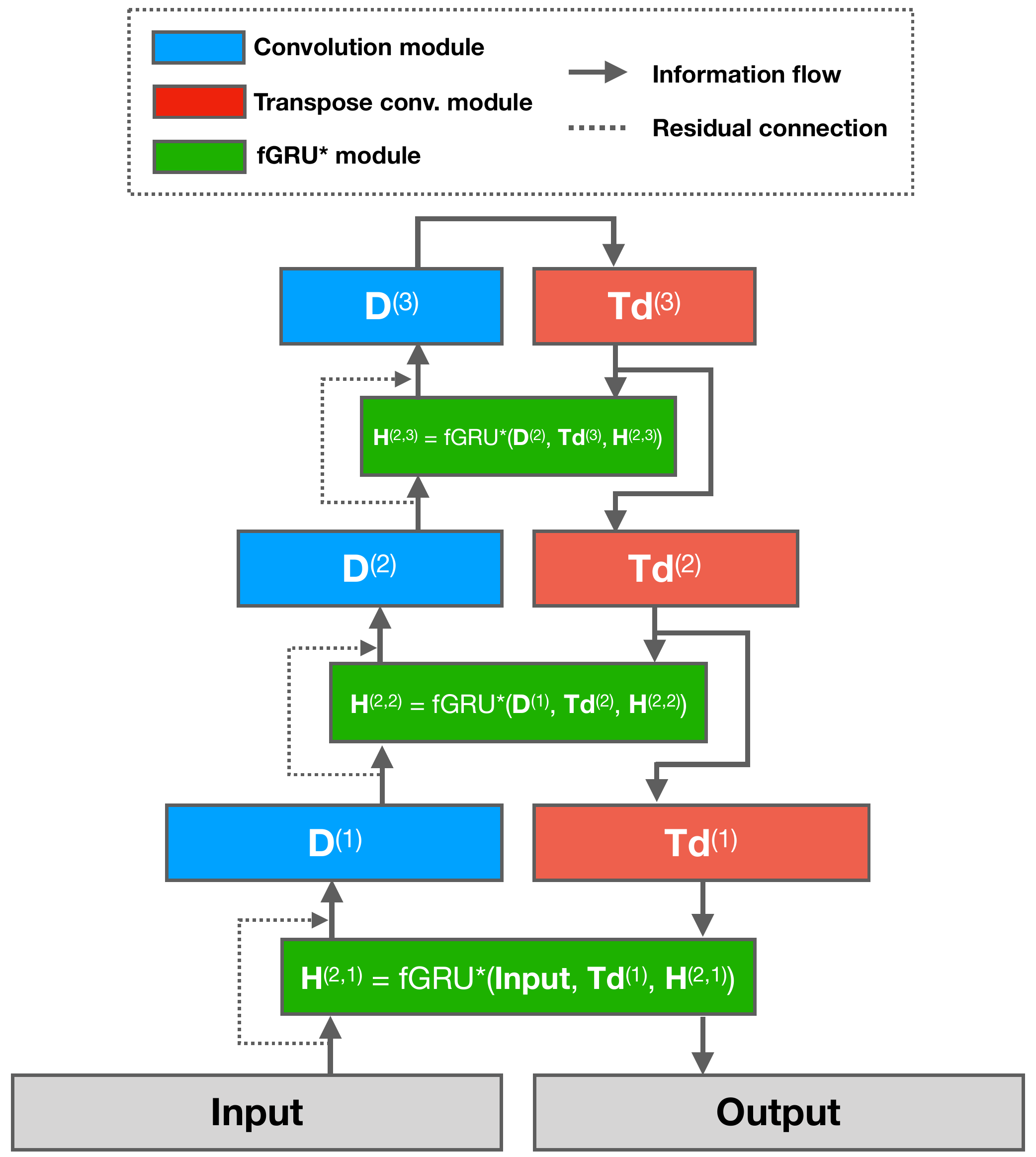}
\end{center}%\vspace{-4mm}
   \caption{A high-level diagram of a FFN-fGRU. Three fGRU* modules are interleaved by three convolutional and three transpose convolutional modules.}
\label{fig:fgru_st_highlevel}
\end{figure}

\paragraph{The FFN-fGRU network} We constructed a three-layer architecture, called the FFN-fGRU, which we pitted against the FFN-CNN in segmenting neurons in the STAR challenge (Fig.~\ref{fig:results1} in the main text and Fig.~\ref{fig:fgru_st_highlevel}, here). This network uses three downsampling and three upsampling steps to combine hierarchical feature processing with horizontal and top-down feedback recurrence at every layer, implemented by fGRU* modules\footnote{We found that increasing model depth led to a significant improvement in performance on this flood-filling segmentation task.}. The shape of horizontal kernels used in the FFN-fGRU model's fGRU* modules (in $\text{height}\times \text{width} \times \text{depth}$) are $7\times 7\times 1$, $5\times 5\times 3$ and $3\times 3\times 3$, respectively. Each convolutional module is a combination of three steps: (1) A combination step which mixes the output of the preceding fGRU module ($\textbf{H}^{(2, l)}$) with a residual bypass ($\textbf{X}$ or $\textbf{D}^{(l-1)}$) from the previous layer (described below); (2) 3-D convolution; and (3) 3-D max pooling and subsampling (Fig.~\ref{fig:fgru_st_highlevel}). 

The combination step of (1) is similar to gain and mix computation in the fGRU* module. This step combines three terms $\textbf{H}^{(2, l)}$, $\textbf{D}^{(l-1)}$, and $\textbf{H}^{(2, l)}\odot\textbf{D}^{(l-1)}$ -- using a trainable scalar weight. The shape of convolution kernels used in the first, second and third convolution modules are $7\times 7\times 1$, $5\times 5\times 1$, and $5\times 5\times 1$, respectively (height/width/depth). The shape of the pooling kernels are $2\times 2\times 1$, $2\times 2\times 2$, and $2\times 2\times 1$, respectively. Subsampling at every stage is performed with the same rate specified by the pooling kernels. Each transpose convolution module uses strided transpose convolution to upsample a high-level feature map to match the shape of a low-level feature map. The stride in each transpose convolutional module is the same as the downsampling rate of the corresponding convolutional module. The shape of convolution kernels in the first, second and third convolution modules are $8\times 8\times 1$, $6\times 6\times 2$, and $6\times 6\times 1$, respectively. The FFN-fGRU network ran for 8 timesteps for every iteration of flood-filling. Like the fGRU discussed in the main text, fGRU* used symmetric weight sharing to reduce its number of free parameters (see Section \ref{sec:membrane_training} in the main text).  % , which is roughly equivalent to 96 steps of convolution between input and output.

%%%%%%%%%%%%%%%%%%%%%%
% Performance table
%%%%%%%%%%%%%%%%%%%%%%
\begin{table}[t]
\begin{tabular}{|l|c|c|c|}
% \hline & mAP & ARAND & Parameters\\ \hline
% FF-Control & 0.44 & X & 60K\\
% 3D U-Net~\cite{Lee2017-ye} & 0.51 & XXX & 1.4M\\
% fGRU-1 & 0.60 & I & 45K\\
% fGRU-2 & \textbf{0.72} & \textbf{I} & 50K\\\hline
% FFN-CNN~\cite{Januszewski2018-rl} & - & XXX & 630K\\
% FFN-fGRU & - & \textbf{I} & 100K\\\hline
\hline & Precision & Recall & Parameters\\ \hline
FFN-CNN~\cite{Januszewski2018-rl} & 0.988 & 0.772 & 630K\\
FFN-CNN-Ctrl & 0.951 & 0.657 & 196K\\
FFN-fGRU & \textbf{0.994} & \textbf{0.853} & 126K\\\hline
\end{tabular}
\linebreak\linebreak%\vspace{-4mm}
\caption{Reference  performance  for  models  after  training  and testing on distinct image splits (50\%/50\%) within the STAR test volume, evaluated by ``patchwise'' precision and recall (higher is better). These metrics are computed as the mean precision and recall of predicted cell masks over 500 iterations of network processing in randomly sampled fields of vision (hence, ``patchwise'').}

\label{tab:ffn_performance_si}%\vspace{-5mm}
\end{table}
%%%%%%%%%%%%%%%%%%
% End performance table
%%%%%%%%%%%%%%%%%%

\paragraph{Training and testing}
Flood-filling networks were trained according to the routines of \cite{Januszewski2018-rl}. The reference implementation of the FFN-CNN was taken from a public repository of those authors (\url{https://www.github.com/google/ffn}). We made two changes to the original configuration that accelerated and improved training. First, the shape of the network's field of view (FOV) changed from the original $33\times33\times17$ voxels to $57\times57\times13$ as it improved the rate of convergence for the FFN-fGRU. We tried this approach with the FFN-CNNs as well; however, we found that its training was more effective with the original field of view, and therefore report results from this configuration. Second, networks were trained with Adam ($1e^{-3}$ learning rate) instead of stochastic gradient descent because it led to a faster convergence. All networks were trained with batches of 16 volumes. Training stopped when validation accuracy, measured in F1 score over a batch of 500 FOVs, stopped improving by more than 3\% over 100K iterations. Validation was performed every 25K training iterations. According to this rule, the FFN-CNN was trained for 475K iterations, the FFN-CNN-Control for 550K iterations, and the FFN-fGRU for 375K iterations. Models were trained on the STAR challenge (see main text), and on a split half of the STAR test set for validation (Table.~\ref{tab:ffn_performance_si}).

\begin{figure*}[t]
\begin{center}
   \includegraphics[width=.9\linewidth]{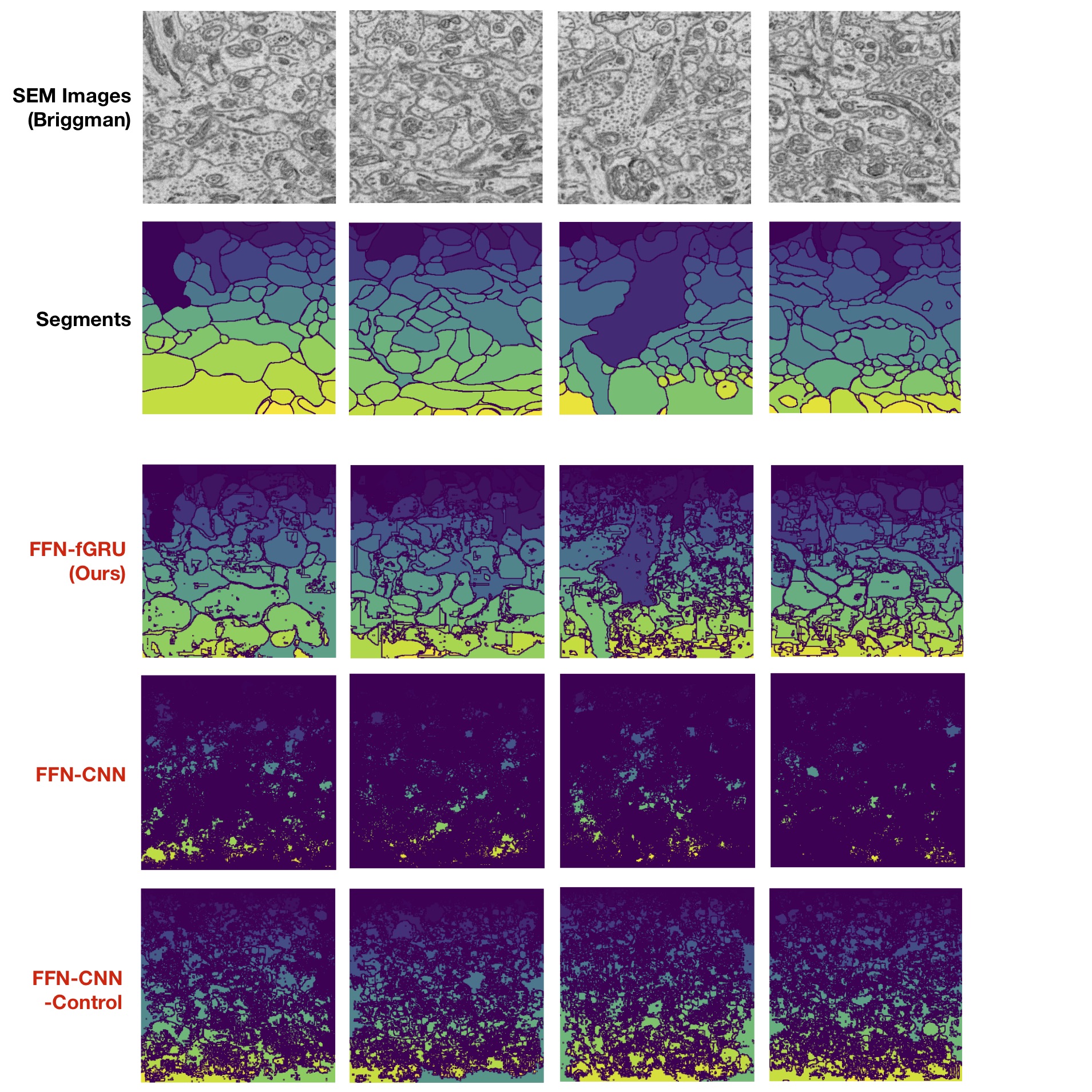}
\end{center}%\vspace{-4mm}
   \caption{Exemplar SEM images from the STAR test dataset (Briggman), their ground-truth neuron annotations, and predicted segmentations from FFN models after training on the STAR train dataset.}
\label{fig:ffn_results_eg}
\end{figure*}

%%%%%%%%%%%%%%%%%%%%%%
% Performance table fib25
%%%%%%%%%%%%%%%%%%%%%%
% \begin{table}[t]
% \centering
% \begin{tabular}{|l|c|c|}
% \hline & mAP & Parameters\\ \hline
% FF-Control & 0.87 & 60K\\
% fGRU-1 & 0.88 & 45K\\
% fGRU-2 & \textbf{0.94} & 50K\\
% 3D U-Net~\cite{Lee2017-ye} & \textbf{0.97} & 1.4M\\\hline
% \end{tabular}
% \linebreak\linebreak%\vspace{-4mm}
% \caption{Reference performance for models after training and testing on distinct image splits (80\%/20\%) of the FIB-25 dataset. Membrane predictions are evaluated by mean Average Precision (mAP; higher is better).}
% \label{tab:fib_performance}%\vspace{-5mm}
% \end{table}
%%%%%%%%%%%%%%%%%%
% End performance table
%%%%%%%%%%%%%%%%%%

%%%%%%%%%%%%%%%%%%%%%%
% Performance table snemi3d
%%%%%%%%%%%%%%%%%%%%%%
% \begin{table}[t]
% \centering
% \begin{tabular}{|l|c|c|}
% \hline & mAP & Parameters\\ \hline
% FF-Control & 0.76 & 60K\\
% fGRU-1 & 0.70 & 45K\\
% fGRU-2 & \textbf{0.89} & 50K\\
% 3D U-Net~\cite{Lee2017-ye} & \textbf{0.91} & 1.4M\\\hline
% \end{tabular}
% \linebreak\linebreak%\vspace{-4mm}
% \caption{Reference performance for models after training and testing on distinct image splits (80\%/20\%) of the FIB-25 dataset. Membrane predictions are evaluated by mean Average Precision (mAP; higher is better).}
% \label{tab:fib_performance}%\vspace{-5mm}
% \end{table}
%%%%%%%%%%%%%%%%%%
% End performance table
%%%%%%%%%%%%%%%%%%

% {
% \small
% \bibliographystyle{ieee}
% \bibliography{Refs,Refs_full} 
% }\flushend

\end{document}